\documentclass{article}
\usepackage{iclr2027_conference,times}

\usepackage{amsmath,amssymb}
\usepackage{graphicx}
\usepackage{multirow}
\usepackage{booktabs}
\usepackage{algorithm}
\usepackage{algpseudocode}
\usepackage{float}
\usepackage{comment}
\usepackage{enumitem}
\floatstyle{ruled}
\restylefloat{algorithm}
\usepackage{placeins}
\usepackage{xcolor}
\usepackage{colortbl}
\usepackage{tcolorbox}
\definecolor{ourframe}{HTML}{7B3FA0}
\definecolor{ourback}{HTML}{FFFFFF}
\definecolor{algstream}{HTML}{E08A1E}
\definecolor{fullcol}{HTML}{FFFFFF}
\definecolor{boxframe}{HTML}{A8871E}
\definecolor{boxback}{HTML}{FFF8DC}
\usepackage{url}
\usepackage[colorlinks=true,linkcolor=blue,citecolor=blue,urlcolor=blue]{hyperref}

\newcommand{\method}{\texttt{JIVEAdapter}}
\newcommand{\R}{\mathbb{R}}

\DeclareMathOperator{\diag}{diag}
\newcommand{\Jm}{\mathbf{J}}          
\renewcommand{\Im}{\mathbf{I}}          
\newcommand{\Wm}{\mathbf{W}}          

\algrenewcommand\algorithmiccomment[1]{\hfill{\color{gray!70}\textit{#1}}}

\newcommand{\iphase}[1]{{\color{gray!85}\textsc{\textbf{#1}:}}\ }
\definecolor{modecol}{HTML}{000000}
\definecolor{reusecol}{HTML}{1F5FBF}

\newcommand{\keybox}[2][boxframe!40!white]{\begin{tcolorbox}[colback=boxback,colframe=#1,boxrule=0.5pt,%
  arc=4pt,left=6pt,right=6pt,top=4pt,bottom=4pt,boxsep=2pt,before skip=5pt,after skip=5pt]%
  \small #2\end{tcolorbox}}

\title{JIVEAdapter: A Multi-Task Additive Low-Rank Adapter
via Joint and Individual Variation Explained (JIVE)}
\author{Sara Abdali \quad Pashmina Cameron\\
Microsoft Applied Sciences Group (ASG)\\
\texttt{\{saraabdali, pcameron\}@microsoft.com}}
\date{}

\begin{document}
\iclrfinalcopy
\maketitle
\lhead{Manuscript under review}\rhead{}

\begin{abstract}
Parameter-efficient fine-tuning adapts pretrained models at a fraction of the cost of
full fine-tuning, yet most low-rank adapters are single-task and represent each weight update
multiplicatively, leaving no explicit account of what is shared across tasks and what is
task-specific. We introduce \method, a multi-task \emph{additive} low-rank adapter inspired by
statistical Joint and Individual Variation Explained (JIVE). \method\ decomposes every weight
update into a Joint structure shared across all tasks plus a per-task Individual structure,
penalizes the Individual structures to be near-orthogonal to the Joint so shared and task-specific signal stay \emph{interpretable} and
separated, and allocates rank adaptively across a shared Joint pool and a per-task Individual
pool. The Joint is learned once, jointly over a task group or incrementally, one task at a
time, then frozen and reused as a prior for new tasks without retraining the shared part. On
GLUE and SuperGLUE with DeBERTaV3-base, \method\ is competitive with strong single-task and
multi-task low-rank baselines at a matched per-task effective rank, without extra modules such as MoE, and when a related held-in task exists its frozen Joint serves a
held-out task by reusing that task's Individual with only a cheap per-direction scale, otherwise training a small new one.
\end{abstract}

\section{Introduction}\label{sec:intro}
Pretrained transformers are now adapted to downstream tasks far more often than they are
trained from scratch, but full fine-tuning stores and updates a separate copy of every weight
per task, which is expensive at scale. Parameter-efficient fine-tuning (PEFT)~\citep{houlsby2019adapter} instead
freezes the pretrained weights and learns a small update; low-rank adapters such as
LoRA~\citep{hu2022lora} and AdaLoRA~\citep{zhang2023adalora} are the dominant methods, adding a
low-rank product to each weight and, in AdaLoRA, allocating a rank budget across layers by
pruning singular values.

Many practical settings involve several related tasks at once, where one model should
exploit what the tasks share while still capturing what makes each task distinct. Standard low-rank
adapters force a poor trade-off: one adapter per task relearns the shared structure from scratch and
transfers nothing across tasks, while a single adapter shared by all tasks suppresses each task's own
variation. Recent multi-task adapters
add an explicit shared component plus per-task
components~\citep{tian2024hydralora,yang2024mtllora,mahabadi2021hyperformer,ponti2023polytropon},
but they represent the update multiplicatively, do not enforce that the shared and
task-specific parts describe non-overlapping variation, and fix or globally allocate the
per-task rank. We propose \method\ (Figure~\ref{fig:stages}), a multi-task additive low-rank adapter inspired by statistical Joint
and Individual Variation Explained (JIVE)~\citep{lock2013jive}. Each weight update is a sum of a
Joint structure shared across all tasks and a per-task Individual structure orthogonal to the
Joint structure, learned on top of a frozen weight with a well-defined initialization. Because the Joint is
estimated from all tasks, a low-resource task inherits it and only needs to fit its small
Individual; and since these structures are explicit and orthogonal, one can
measure how much each task relies on Joint versus Individual structure. Our contributions are:
\begin{itemize}
  \item Additive, orthogonal shared/individual split. Unlike multiplicative adapters,
  \method\ writes the update as the sum of the Joint and the Individual and constrains the
  Individual to be orthogonal to the Joint, borrowing the JIVE
  principle so the shared Joint captures overlapping and each Individual non-overlapping
  variation.
  \item Joint and Individual pool adaptive rank. Rank is allocated adaptively across a shared Joint pool
  and a per-task Individual pool, rather than across layers as in AdaLoRA, so the shared
  structure is not starved by many per-task structures.
  \item A reusable frozen Joint. The Joint is learned once, jointly over a task group
  or incrementally, one task at a time, then frozen and reused as a prior for a later task
  without retraining the shared part, either by reusing a similar task's Individual or by
  training a small new one.
  \item Empirical validation. On GLUE and SuperGLUE with DeBERTaV3-base, \method\
  is competitive with strong single-task and fully-shared multi-task low-rank baselines at a matched per-task effective rank, and
  analysis confirms the Joint represents shared structure while Individuals represent
  task-specific variation, exposing the
  underlying task structure.
\end{itemize}

\section{Related Work}\label{sec:related}
\paragraph{Single-task low-rank adaptation.}
Low-rank adapters add a trainable update to the frozen weights instead of tuning all parameters. The single-task counterparts of \method\ are LoRA~\citep{hu2022lora}, a fixed-rank update, and AdaLoRA~\citep{zhang2023adalora}, which reparametrizes it as an SVD triplet and prunes singular values under a global budget. AdaLoRA is the closest single-task counterpart; unlike it, \method\ parametrizes each structure as low rank using SVD while decomposing the update with JIVE~\citep{lock2013jive} rather than a naive SVD, giving a sum of task-specific orthogonal structures with rank allocated per structure, not per layer.

\paragraph{Multi-task low-rank adaptation.}
The multi-task counterparts, MTL-LoRA~\citep{yang2024mtllora} and HydraLoRA~\citep{tian2024hydralora}, add task-specific factors on a shared down-projection, and several such methods rely on mixture-of-experts (MoE) routing over multiple experts~\citep{tian2024hydralora,ning2025mode}. \method, however, splits capacity into an interpretable, reusable additive sum of a Joint and an Individual component under a JIVE orthogonality constraint, allocates per-task rank adaptively, and leaves the backbone unchanged, without any MoE routing.

\paragraph{How \method\ differs.}
\method\ is the only method that combines an explicit shared and task-specific split, orthogonal Joint and Individual subspaces, per-structure adaptive rank, and a frozen Joint that transfers to new tasks, reused for unseen tasks either by similarity or by training a small new Individual, all as a low-rank update that leaves the backbone unchanged. For extended related work see Appendix~\ref{sec:apprelated}.

\section{Method}\label{sec:method}
\keybox{\textbf{\color{boxframe}Motivation:} We propose an \textbf{interpretable} adaptation that models the \textbf{individual}
characteristics of a task while relating it to similar tasks through an \textbf{orthogonal Joint}
structure: a shared \textbf{Joint} that captures
what tasks have in common and exposes how they relate to one another. This is exactly why we
adopt \textbf{JIVE}: it gives a principled way to separate shared from
individual variation, letting us represent related tasks as separate subspaces while still
relating them through a common shared space. Moreover, we propose frameworks that make our adapter applicable not
only to \textbf{single-task} and \textbf{multi-task} scenarios, but also when a pool of tasks already
exists or \textbf{grows over time (incremental)}.}

\begin{figure}[t]
  \centering
  \includegraphics[width=\linewidth]{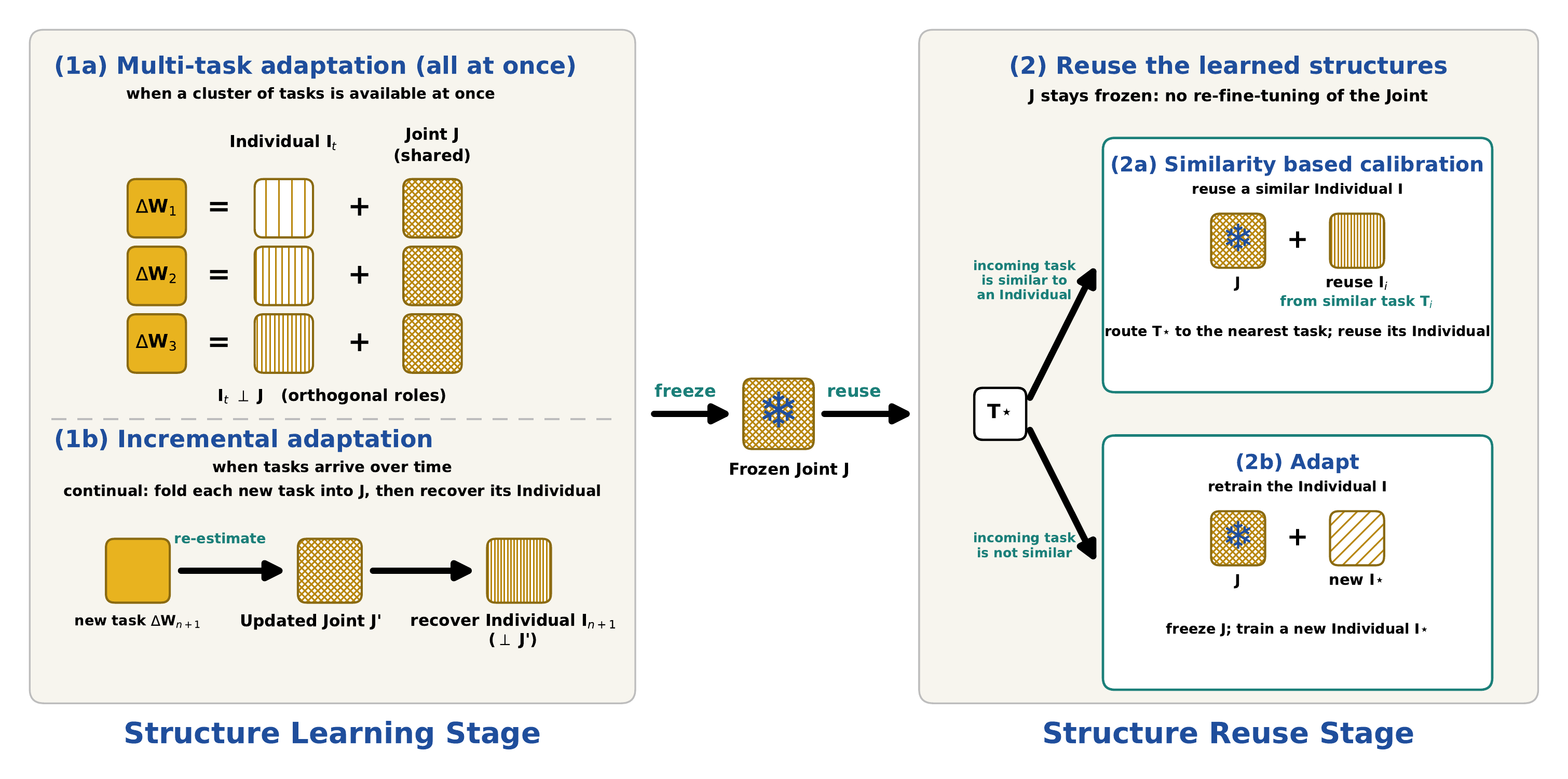}
  \caption{\method\ in two stages. The Structure Learning Stage builds a shared Joint $\Jm$ with
  orthogonal per-task Individuals $\Im_t\perp\Jm$ and freezes it into $\Jm^\star$, by
  (1a) multi-task adaptation or (1b) incremental adaptation
  (add each new task in, then recover its Individual). The Structure Reuse Stage serves a new task
  $T^\star$ without retraining $\Jm^\star$: (2a) {\color{reusecol}\textsc{similarity}} reuses a
  resembling task's Individual, else (2b) {\color{reusecol}\textsc{Adapt}} trains a new Individual
  $\Im^\star$.}
  \label{fig:stages}
\end{figure}

\subsection{Formulation}\label{sec:setup}
JIVE (Joint and Individual Variation Explained)~\citep{lock2013jive}, later refined to measure the
shared structure through principal angles between subspaces~\citep{feng2018ajive}, is a statistical
decomposition that, given several related data sources, splits each source into a Joint
structure shared across all sources and a source-specific Individual structure, with each
Individual constrained to be orthogonal to the Joint so that the two extract non-overlapping
variation. We leverage this in parameter-efficient adaptation through gradient-based learning rather than a linear least-squares fit; each per-task weight update is the
sum of a Joint update shared by all tasks and a task-specific Individual update, and we impose the
same orthogonality so the shared and task-specific parts span non-overlapping subspaces.

\paragraph{Formulation.}
Let $\Wm\in\R^{\text{out}\times\text{in}}$ be a frozen weight; only the Joint and Individual factors below are trained. We attach a trainable, additive update
\begin{equation}
  \Delta\Wm_t \;=\; \Jm \;+\; \Im_t, \qquad \Jm,\ \Im_t\in\R^{\text{out}\times\text{in}},
\end{equation}
where $\Jm$ is shared by all $T$ tasks and $\Im_t$ is the Individual update of task
$t\in\{1,\dots,T\}$. The adapted layer output for a sample belonging to task $t$ is
\begin{equation}
  y \;=\; x\Wm^\top \;+\; \mathrm{drop}(x)\,\Jm^\top \;+\; \mathrm{drop}(x)\,\Im_t^\top .
  \label{eq:forward}
\end{equation}

\paragraph{Parametrizing each structure.}
Each structure is a low-rank SVD triplet $U\,\diag(E)\,V$:
\begin{align}
  \Jm      &= \tfrac{\alpha_J}{r_J+\varepsilon}\; U_J\,(E_J\odot V_J), &
  V_J&\in\R^{r_J\times \text{in}},\ E_J\in\R^{r_J\times1},\ U_J\in\R^{\text{out}\times r_J},\\
  \Im_t    &= \tfrac{\alpha_I}{r_I+\varepsilon}\; U_{I,t}\,(E_{I,t}\odot V_{I,t}), &
  V_{I,t}&\in\R^{r_I\times \text{in}},\ E_{I,t}\in\R^{r_I\times1},\ U_{I,t}\in\R^{\text{out}\times r_I}.
\end{align}
with the scaling factor of $\tfrac{\alpha}{r+\varepsilon}$, where $\alpha_J,\alpha_I$ are fixed
scaling hyperparameters and regularizing by the rank $r$ keeps the effective magnitude of each
update roughly constant as $r$ varies, so rank and $\alpha$ can be tuned independently
($\varepsilon$ guards against $r{=}0$). Each Individual factor carries an extra leading task axis of size $T$, while the Joint factors
are single shared tensors.

\subsection{Proposed \method}\label{sec:proposed}
\method\ has two stages (Fig.~\ref{fig:stages}). The \textbf{Structure Learning Stage} builds a single
shared Joint $\Jm$ per layer plus one small per-task Individual $\Im_t$,
constrained to be near-orthogonal so that shared and task-specific structure are largely separated; the
Joint is built either by \textbf{(1a)} multi-task training on a related group of tasks (all
at once) or by \textbf{(1b)} incremental adaptation, adding each new task into the Joint and recovering its
Individual. The \textbf{Structure Reuse Stage} then reuses the frozen Joint as a reusable prior for
a new task that arrives later, never retraining the shared part, by \textbf{(2a)} reusing a
similar task's Individual when the new task resembles a trained one or \textbf{(2b)} training
a new Individual otherwise.

\subsubsection{Structure Learning Stage: Extracting the Joint Structure}\label{sec:stage1}

\paragraph{(1a) Multi-Task Training: Shared Joint, Per-Task Individuals}\label{sec:multitask}
We aim to learn several tasks simultaneously, namely a single shared Joint
plus one Individual per task, so that at deployment the Joint is fixed and only the small
per-task Individual is substituted in. This matches canonical JIVE, where the joint structure
is estimated jointly from all tasks rather than any one of them.
A batch may contain samples from several tasks. Each layer receives a per-sample task-id vector and routes each sample to its own Individual while the shared Joint applies to all samples (Alg.~\ref{alg:forward}). Grouping by the distinct tasks in the batch keeps the cost to one small matmul per active task, not per sample.

\begin{algorithm}[h]
\caption{Forward for one \method\ layer (single or multitask)}\label{alg:forward}
\fontsize{6.2}{7.3}\selectfont
\begin{algorithmic}[1]
\Require input $x$ (batch $\times$ \dots $\times$ in), per-sample task ids, frozen $\Wm$
\Ensure adapted output $y$
\State $y \gets x\Wm^\top$ \Comment{frozen base}
\State $\Jm \gets \tfrac{\alpha_J}{r_J+\varepsilon} U_J(E_J\odot V_J)$
\State $y \gets y + \mathrm{drop}(x)\,\Jm^\top$ \Comment{shared Joint, all samples}
\If{task ids is None \textbf{or} $T=1$}
  \State \Return $y + \mathrm{drop}(x)\,\Im_1^\top$ \Comment{single-task: one Individual, no routing}
\EndIf
\Statex \Comment{multi-task ($T>1$): route each sample to its own Individual below}
\State $\mathrm{ind}\gets \mathbf{0}$ \Comment{output-shaped Individual buffer, zeros like $y$}
\For{$t \in \mathrm{unique}(\text{task ids})$}
  \State $m \gets (\text{task ids}=t)$ \Comment{boolean mask over batch}
  \State $U_{I,t}\gets(\mathbb{I}-U_JU_J^\top)U_{I,t}$ \Comment{$\Im_t\perp\Jm$}
  \State $\Im_t \gets \tfrac{\alpha_I}{r_I+\varepsilon} U_{I,t}(E_{I,t}\odot V_{I,t})$
  \State $\mathrm{ind}[m]\gets \mathrm{drop}(x[m])\,\Im_t^\top$
\EndFor
\State \Return $y + \mathrm{ind}$
\end{algorithmic}
\end{algorithm}

Under Eq.~\eqref{eq:forward}, in a mixed batch the shared Joint receives gradient from
every task at each step, while each Individual receives gradient only from its own
samples. Thus the Joint converges to structure common to all tasks (the SGD analogue of
estimating $\mathbf J$ from the concatenation of all blocks), and the Individuals absorb
task-specific residuals. This was verified empirically (Joint gradient non-zero from any
task; Individual $t$ gradient non-zero only when task $t$ is present).
Mixed batches keep the shared Joint stable (one averaged gradient over all tasks per
step); alternating per-task batches makes the Joint oscillate toward whichever task it last
saw. A compromise is to accumulate gradients over homogeneous per-task micro-batches, then take a single optimizer step.

\paragraph{Orthogonal Regularization terms in loss}\label{sec:loss}
The training objective is the task loss plus $\lambda\,\mathcal{R}$, where $\mathcal{R}$
enforces orthonormal SVD factors for the Joint and every Individual across all \method\ layers,
with Individual$\perp$Joint handled by the forward projection,
\begin{equation}
\mathcal{R}=\sum_{\text{layers}}\!\Bigl(\|V_JV_J^\top-\mathbb{I}\|_F+\|U_J^\top U_J-\mathbb{I}\|_F
+\!\sum_{t}\|V_{I,t}V_{I,t}^\top-\mathbb{I}\|_F+\|U_{I,t}^\top U_{I,t}-\mathbb{I}\|_F\Bigr).
\label{eq:ortho-reg}
\end{equation}

\paragraph{Imposing orthogonality in the forward pass}\label{sec:orthproj}
We keep the Joint's output factor $U_J$ (approximately) column-orthonormal via the
orthogonality regularizer $\mathcal{R}$ (Eq.~\ref{eq:ortho-reg}), where the columns of $U_J$
span the update's output space. Given that, we can remove any overlap between an
Individual and the Joint directly in the forward pass by projecting the Individual's output
factor onto the orthogonal complement of the Joint's column space:
\begin{equation}
  \widetilde{U}_{I,t} \;=\; \bigl(\mathbb{I}-U_JU_J^\top\bigr)\,U_{I,t},
  \qquad
  \Im_t=\tfrac{\alpha_I}{r_I+\varepsilon}\,\widetilde{U}_{I,t}\,(E_{I,t}\odot V_{I,t}).
  \label{eq:proj}
\end{equation}
where $\mathbb{I}$ is the identity matrix. The projection is exact when $U_J^\top U_J=\mathbb{I}$, enforced by the regularizer.

\paragraph{Adaptive Rank Allocation}\label{sec:alloc}\label{sec:permrank}
We allocate rank per structure with two strategies. The first is an importance-based
allocator adapted from AdaLoRA, which prunes singular values under a decaying budget during
training. The second is the budget-free permutation-test estimator native to statistical
JIVE~\citep{lock2013jive}, adapted to weight space and run once before training: it estimates the
shared Joint and per-task Individual ranks directly from data, with no importance schedule and
no pruning. In both, the shared Joint pool and the per-task Individual pool are budgeted
separately, so many per-task structures cannot starve the single shared Joint. Full details of
both algorithms are deferred to Appendix~\ref{sec:apprankalgs}
(Alg.~\ref{alg:alloc} and Alg.~\ref{alg:permrank}).

\paragraph{Merging at inference.}
Merging is a LoRA-style inference trick that folds a fixed adapter update $\Delta\Wm$ into
the frozen backbone weight $\Wm$, so a merged layer runs one matmul instead of two. It is
purely an efficiency rewrite that leaves the output unchanged, and it is enabled only for a
single task ($T{=}1$), where the one update $\Jm+\Im_1$ can be folded into $\Wm$ once. With
several task-specific Individuals there is no single update to merge, since folding one task's
$\Im_t$ into the shared $\Wm$ would corrupt every other task, so merging is automatically
disabled when $T>1$. Each layer then keeps the frozen $\Wm$, adds the shared Joint $\Jm$ to
every sample, and routes each sample to its own Individual $\Im_t$ on the fly
(Alg.~\ref{alg:forward}). Recovering merging under $T>1$ would need a separate merged weight
per task, i.e.\ $T$ full weight matrices in every layer and a matmul per task, so we keep the
frozen $\Wm$ plus cheap low-rank additions instead.
The routed forward, the orthogonal regularizer, and the per-structure allocator compose into a
single optimizer step, with a separate head per task because the label spaces~differ.

\paragraph{(1b) Incremental Task Adaptation}\label{sec:incremental}
As an alternative to learning the Joint from all tasks at once (1a), \method\ can build it
incrementally: the Joint moves to reflect each newly arrived task. Given $n$ tasks with a shared Joint $\Jm$ and Individuals
$\{\Im_t\}$, and one incoming task, we (i) reestimate the Joint over all $n{+}1$ tasks
warm-started from the current $\Jm$, then (ii) recover the incoming Individual (and refresh
the existing ones) as the rank-$r_I$ residual orthogonal to the updated Joint. Given
$Q_J=\mathrm{col}(U_J)$ for the shared output subspace and $P_{Q}= QQ^\top$ for the projector
onto a subspace $Q$, the estimator alternates two closed-form steps: a joint step that
fits the single shared matrix $\Jm'=P_{Q_J'}\overline{\Delta\Wm}$ best explaining the weighted-mean
update $\overline{\Delta\Wm}$ inside the reestimated shared subspace $Q_J'$, and an individual
step $\Im_t=\mathrm{trunc}_{r_I}\big((\mathbb{I}-P_{Q_J'})\Delta\Wm_t\big)$ that takes each task's
residual orthogonal to $\Jm'$ (Alg.~\ref{alg:incjive}). The special case that keeps
$\Jm$ fixed recovers exactly the frozen-Joint estimate
$\Im_{n+1}=\mathrm{trunc}_{r_I}\big((\mathbb{I}-P_{Q_J})\Delta\Wm^{\star}_{n+1}\big)$; letting
$Q_J'$ rotate/grow lets a sufficiently informative new task refine the shared
structure, and the per-task Individuals stay orthogonal to it by construction.

\begin{algorithm}[h]
\caption{Incremental JIVE: fold a new task into the Joint and recover its Individual}\label{alg:incjive}
\fontsize{6.2}{7.3}\selectfont
\begin{algorithmic}[1]
\Require frozen base weight $\Wm$; current joint $\Jm{=}U_J\diag(E_J)V_J$ with subspace $Q_J{=}\mathrm{orth}(U_J)$ and rank $r_J$; existing individuals $\{\Im_t\}_{t=1}^{n}$; incoming task data $D_{n+1}$; individual rank $r_I$; iterations $K$; share threshold $\tau$; growth cap $\kappa$; per-task delta weights $\{\gamma_t\}$
\Ensure updated joint $\Jm'$; incoming individual $\Im_{n+1}$; refreshed $\{\Im_t\}$

\State $\Delta\Wm_t\gets \Jm+\Im_t$,\ \ $t\le n$ \Comment{construct $\Delta\Wm$ per existing task}
\State $\Im_{n+1}^{(0)}\gets\arg\min_{\mathrm{rank}(\Im)\le r_I}\mathcal L(\Wm{+}\Jm{+}\Im;\,D_{n+1})$,\ \ $\Delta\Wm_{n+1}\gets\Jm+\Im_{n+1}^{(0)}$ \Comment{incoming task update}
\State $\widehat\Sigma_t\gets \Delta\Wm_t\Delta\Wm_t^\top/\|\Delta\Wm_t\|_F^2,\quad t\le n{+}1$

\For{$k=1,\dots,K$}
  \State $C\gets \textstyle\sum_{t=1}^{n+1}\gamma_t\,\widehat\Sigma_t$;\quad
  $Q_J'\gets \mathrm{eig}_{r_J}(C)$ \Comment{top-$r_J$ eigenvectors, warm-started from $Q_J$}
  \State $Q_J'\gets\{\,u\in Q_J' \,:\, \min_{t\le n+1}\, u^\top\widehat\Sigma_t\,u\ \ge\ \tau\,\}$ \Comment{keep directions shared by all tasks}
  \State $\overline{\Delta\Wm}\gets \big(\textstyle\sum_t \gamma_t\Delta\Wm_t\big)\big/\textstyle\sum_t \gamma_t$ \Comment{weighted-mean update}
  \State $P_{Q_J'}\gets Q_J'Q_J'^\top$;\quad $\Jm'\gets P_{Q_J'}\,\overline{\Delta\Wm}$ \Comment{new Joint: averaged update projected onto the shared subspace}
  \State $\Im_t\gets \mathrm{trunc}_{r_I}\!\big((\mathbb{I}-P_{Q_J'})\,\Delta\Wm_t\big),\quad t\le n{+}1$ \Comment{recover each Individual off the Joint}
  \State $Q_J\gets Q_J'$ \Comment{stop when the shared subspace stops moving: $d_{\mathrm{Gr}}(Q_J,Q_J')<\epsilon$}
\EndFor

\State $\Jm'=U_J'\diag(E_J')V_J'$ (truncated SVD, rank $r_J$);\quad append $\Im_{n+1}$ to the pool
\State \Return $\Jm',\ \Im_{n+1},\ \{\Im_t\}$ \Comment{$Q_J'^\top U_{I,t}=0$}
\end{algorithmic}
\end{algorithm}

\subsubsection{Structure Reuse Stage}\label{sec:sharing}

\label{sec:reuse}
Given a shared Joint $\Jm$ extracted from a group of tasks and a new task
$T^\star$, we freeze both the pretrained weights $\Wm$ and the
shared Joint $\Jm$ and reuse $\Jm$ as a prior, learning at most a small task-specific
Individual instead of adapting from scratch. Algorithm~\ref{alg:reuse} makes this precise
through two modes, neither of which retrains the shared part.

\textbf{(2a) {\color{modecol}\textsc{Reuse similar task Individual}}.} Inspired by the notion that a
  task is characterized by the gradients it produces on a frozen backbone~\citep{achille2019task2vec} and by the SVD gradient-subspace construction of Gradient Projection Memory (GPM)~\citep{saha2021gpm}, which we adapt from activations to weight-space
  updates, we identify the prior task most similar to $T^\star$ and reuse its Individual, retraining
  none of its basis and learning only one scalar gain per rank direction. We summarize $T^\star$ by
  a signature $G_{T^\star}$: the top-$r_I$ left singular vectors of its frozen-model gradient
  averaged over $B$ minibatches ($r_I$ the adapter rank, $\mathcal L$ the loss at the frozen
  $\Wm+\Jm$ with the task's Individual switched off),
  \begin{equation}
    \bar\nabla_{T^\star}=\frac1B\sum_{b=1}^{B}\nabla_{\Wm}\mathcal L(x_b,y_b;\Wm{+}\Jm)=U\Sigma V^\top,
    \qquad
    G_{T^\star}=U_{:,\,1:r_I},
    \label{eq:tasksig}
  \end{equation}
  an orthonormal basis ($G_{T^\star}^\top G_{T^\star}=\mathbb{I}_{r_I}$) for the subspace carrying
  most of $T^\star$'s gradient energy. Computing an analogous signature $G_t$ for every prior task,
  we route $T^\star$ to its closest task $T_c$ by principal-angle overlap~\citep{bjorck1973angles} of the two
  signatures,
  \begin{equation}
    T_c=\arg\max_{t\le n}\,\mathrm{overlap}(G_{T^\star},G_t),
    \qquad
    \mathrm{overlap}(G,G')=\frac1{r_I}\|G^\top G'\|_F^2=\frac1{r_I}\sum_{i=1}^{r_I}\cos^2\theta_i,
    \label{eq:overlap}
  \end{equation}
  where the $\cos\theta_i$ are the singular values of $G_{T^\star}^\top G_t$, so the score
  $\in[0,1]$ and is $1$ when subspaces overlap.
  We route on the Individual alone, not on $\Wm+\Jm+\Im$: the frozen $\Wm$ is shared across tasks
  and the Joint $\Jm$ is common to every task (and already reused by $T^\star$), so both inject
  task-independent mass that would dominate and wash out the score, whereas each $\Im_t\perp\Jm$
  isolates exactly the task-specific, transferable part.
  We take the routed task's Individual $\Im_{T_c}$, which is already orthogonal to the frozen $\Jm$ by
  construction, and reuse it directly ($\Im_{T^\star}=\Im_{T_c}$), keeping its two basis matrices
  $(U_{I_{T_c}},V_{I_{T_c}})$ frozen. We then train only the $r_I$ diagonal gains $E_{T^\star}$ (one
  importance scaler per direction) and a fresh
  classification head: freezing the borrowed basis keeps $T^\star$ in the \emph{same} subspace as its
  neighbour $T_c$, while learning $E_{T^\star}$ re-weights how strongly each direction of that
  subspace is used, adapting the shared structure to $T^\star$ at negligible cost.

\textbf{(2b) {\color{modecol}\textsc{Adapt}}.} Freeze $\Wm+\Jm$ and train only a small new Individual
  $\Im_{T^\star}$ (rank $r_I$), so the new task learns exactly the part the Joint
  does not already explain.
How
much of the new task already lives in the Joint is quantified by the shared-energy
ratio~\citep{lock2013jive,saha2021gpm}, the fraction of joint variation explained in the sense of linear JIVE,
$\eta_{T^\star}=\|U_JU_J^\top\widehat{\Delta\Wm}\|_F/\|\widehat{\Delta\Wm}\|_F$, where $\widehat{\Delta\Wm}$
is a reference single-task update $\Delta\Wm^{\star}_{T^\star}$ (single-task LoRA or full
fine-tuning) when available, else the task signature $G_{T^\star}$. Decomposing it against
the frozen Joint,
\[
  \underbrace{\Jm\text{-part}=U_JU_J^\top\,\widehat{\Delta\Wm}}_{\text{explained by the shared Joint}},
  \qquad
  \underbrace{\widehat{\Im}_{T^\star}=(\mathbb{I}-U_JU_J^\top)\,\widehat{\Delta\Wm}}_{\text{residual }=\text{ recovered Individual}} ,
\]
gives a high $\eta_{T^\star}$ exactly when the shared structure carries most of the task, so
similarity based calibration already transfers well; the orthogonal residual
$\widehat{\Im}_{T^\star}$ is the part the new task must still learn, and plugging it (or a
small learned $\Im_{T^\star}$) on top of the frozen $\Wm+\Jm$ should reconstruct $T^\star$'s
accuracy at a fraction of the trainable parameters and data.

\begin{algorithm}[t]
\caption{Reusing a frozen Joint for a new task $T^\star$}\label{alg:reuse}
\fontsize{6.2}{7.3}\selectfont
\begin{algorithmic}[1]
\Require frozen $\Wm$; frozen Joint $\Jm{=}U_J\diag(E_J)V_J$, $U_J^\top U_J{=}\mathbb{I}$; individuals $\{\Im_t\}_{t=1}^{n}$; held-in task signatures $\{G_t\}_{t=1}^{n}$, $G_t{=}\mathrm{U}_{r_I}\!\big(\nabla_{\Wm}\mathcal L(\Wm{+}\Jm;\,B_t)\big)$; data $D_{T^\star}$; task loss $\mathcal L(\cdot;D)$; rank $r_I$; mode $\in\{\text{similarity based calibration},\ \text{adapt}\}$
\Ensure individual $\Im_{T^\star}$; shared-energy ratio $\eta_{T^\star}$
\If{mode $=$ similarity based calibration}
  \State $G_{T^\star}\gets \mathrm{U}_{r_I}\!\big(\nabla_{\Wm}\mathcal L(\Wm{+}\Jm;\,B)\big),\ \ B\subset D_{T^\star}$ \Comment{task signature: top-$r_I$ left singular vectors}
  \State $T_c\gets \arg\max_{t\le n}\ \mathrm{overlap}(G_{T^\star},G_t)$ \Comment{hard nearest-task routing on signatures, Eq.~\eqref{eq:overlap}}
  \State $\Im_{T^\star}\gets \Im_{T_c} \;=\; U_{I_{T_c}}\diag(E_{T_c})V_{I_{T_c}}$ \Comment{reuse nearest Individual directly (already $\perp\Jm$)}
  \State $E_{T^\star}\gets \arg\min_{E}\ \mathcal L\big(\Wm{+}\Jm{+}U_{I_{T_c}}\diag(E)V_{I_{T_c}};\,D_{T^\star}\big)$ \Comment{scale-only; borrowed basis $U_{I_{T_c}},V_{I_{T_c}}$ fixed}
\Else \Comment{adapt}
  \State $\Im_{T^\star}\gets \arg\min_{\mathrm{rank}(\Im)\le r_I}\ \mathcal L(\Wm{+}\Jm{+}\Im;\,D_{T^\star}),\quad \Im\!\leftarrow\!\mathbf 0$ \Comment{train a fresh Individual}
\EndIf
\State $\widehat{\Delta\Wm}\gets \Delta\Wm^{\star}_{T^\star}$ \textbf{if} a reference update is available \textbf{else} $G_{T^\star}$ \Comment{reference $\Delta\Wm^{\star}_{T^\star}$: a separately trained single-task LoRA or full fine-tune on $T^\star$; used only for the diagnostic $\eta$, not for reuse}
\State $\eta_{T^\star}\gets \|U_JU_J^\top\widehat{\Delta\Wm}\|_F\,/\,\|\widehat{\Delta\Wm}\|_F$ \Comment{diagnostic: fraction of the task already in the Joint}
\State \Return $\Im_{T^\star},\ \eta_{T^\star}$ \Comment{$U_J^\top U_{I_{T^\star}}=0$}
\end{algorithmic}
\end{algorithm}

\section{Experiments}\label{sec:bench}
We adopt the AdaLoRA experimental setting~\citep{zhang2023adalora} as our common evaluation framework so that \method\ and every
baseline are compared on identical data, backbone, and parameter budget. We evaluate on
the GLUE benchmark~\citep{wang2019glue}, reporting each task's
standard metric, and additionally on the
SuperGLUE benchmark~\citep{wang2019superglue}. For the structure-reuse study we further use the MNLI
corpus~\citep{williams2018mnli}, treating its five annotated genres (fiction, government, slate,
telephone, and travel) as a cross-genre pool. We use
DeBERTaV3-base~\citep{he2023debertav3} as the backbone, with adapters on the query,
key, value, intermediate, layer-output, and attention-output projections (identical to the
AdaLoRA setup). We match the per-task effective rank across methods rather than the raw parameter count, which differs by structure;
the full setup (baseline taxonomy, model parameters and invariants and
budget split), including a single fixed random seed shared by all methods and every hyperparameter, is
detailed in Appendix~\ref{sec:appsetup} for reproducibility.

\paragraph{Baselines.}\label{sec:baselines}
We run two single-task low-rank adapters, LoRA~\citep{hu2022lora} and AdaLoRA~\citep{zhang2023adalora}, two structured multi-task LoRA methods, HydraLoRA~\citep{tian2024hydralora} and MTL-LoRA~\citep{yang2024mtllora}, and \method; their capacity structure is compared in Table~\ref{tab:baselines} (Appendix~\ref{sec:appsetup}). We additionally include two sequential continual-learning references in the incremental study, a single shared adapter fine-tuned across the task stream and full fine-tuning of the entire backbone, both without any frozen shared structure (Table~\ref{tab:incremental}). Methods that add auxiliary modules, notably mixture-of-experts (MoE) routing such as MoDE, are not directly comparable, as those modules enlarge the parameter and compute budget beyond a single low-rank update; \method\ uses none, and of that family we include only HydraLoRA, the closest to ours.

\paragraph{Case studies.} Following Figure~\ref{fig:stages}, we evaluate \method\ along six axes grouped by the method's two stages: a single-task setting, then the Structure Learning Stage (building the Joint) and the Structure Reuse Stage, followed by two analysis axes.

\begin{enumerate}[nosep,leftmargin=*]
\item \textbf{Single-task} setting ($T{=}1$) trains one task at a time, where \method\ reduces to Joint\,$+$\,one Individual, comparing our orthogonal additive parametrization and per-structure allocator to the multiplicative low-rank updates of LoRA/AdaLoRA, whose premise follows the intrinsic dimensionality of fine-tuning~\citep{aghajanyan2021intrinsic} (Table~\ref{tab:single_glue}).

\item \textbf{Multi-task all-at-once training} (Stage 1a, Structure Learning Stage) trains all tasks jointly with mixed-task batches, one shared Joint and one Individual per task, on GLUE and SuperGLUE (Table~\ref{tab:multi}); at inference the Joint is fixed and only the small task Individual is selected.

\item \textbf{Multi-task incremental training} (Stage 1b, Structure Learning Stage) learns a base Joint from a task cluster and folds each new task in one at a time by refining both the shared Joint and its own Individual following Algorithm~\ref{alg:incjive}, versus two sequential fine-tuning references (a single shared adapter and full backbone fine-tuning) and all-at-once joint training (Table~\ref{tab:incremental}).

\item \textbf{Similarity (2a) and adapt (2b) structure reuse} serves a held-out task from a frozen Joint, either routing to the highest-overlap held-in Individual (2a) or training a small new one (2b), evaluated leave-one-out on the SuperGLUE cross-task and MNLI cross-genre pools (Figure~\ref{fig:reuse}).

\item \textbf{Ablation of the rank-estimation strategy} compares the importance-based low-rank per-structure allocator against the permutation-test rank estimator of JIVE (Appendix~\ref{sec:apprankalgs}).

\item \textbf{Interpretability analysis} of what the Joint and Individuals learn, showing each Individual settles near-orthogonal to the shared Joint and that task-signature similarity routes each held-out task to its most related held-in Individual (Appendix~\ref{sec:appexp}).
\end{enumerate}

\subsection{Results}\label{sec:results}

\subsubsection{Single-Task Comparison ($T{=}1$)}\label{sec:res-found}
\paragraph{Single-task.} \method\ attains the best single-task average on both
benchmarks. On GLUE it reaches $89.44$, ahead of our reproduced AdaLoRA reference
($89.14$) and LoRA ($87.90$); on SuperGLUE it reaches $78.65$, ahead of LoRA ($78.41$)
and AdaLoRA ($77.21$). It leads on five of eight GLUE tasks and on CB, WiC, and ReCoRD,
showing the additive parametrization and per-structure allocator do not cost accuracy
before any multi-task sharing.

\begin{table}[h]\centering\scriptsize
\caption{Single-task evaluation (best per task), a separate adapter trained with no sharing;
\method\ stays competitive even without sharing. \#P is trainable parameters per task head.}
\label{tab:single_glue}
\renewcommand{\arraystretch}{1.22}
\setlength{\aboverulesep}{0.05ex}
\setlength{\belowrulesep}{0.05ex}
\resizebox{\textwidth}{!}{%
\begin{tabular}{clcccccccccc}
\toprule
& Method & MNLI {\tiny(acc)} & QQP {\tiny(acc/F1)} & QNLI {\tiny(acc)} & SST-2 {\tiny(acc)} & CoLA {\tiny(Mcc)} & RTE {\tiny(acc)} & MRPC {\tiny(acc/F1)} & STS-B {\tiny(P/S)} & Avg & \#P {\tiny(M)}\\
\cmidrule(lr){2-12}
& LoRA & 89.52 & \textbf{92.21}/\textbf{89.67} & 93.79 & 94.72 & 70.44 & 81.59 & 89.22/92.14 & \textbf{91.73}/\underline{91.69} & 87.90 & \textbf{1.33}\\
\rowcolor{fullcol}& AdaLoRA ($\bar r{=}8$, repro) & \underline{90.43} & \underline{91.97}/\underline{89.44} & \underline{94.33} & \underline{95.41} & \textbf{71.91} & \underline{87.36} & \underline{90.20}/\underline{92.91} & \underline{91.58}/91.49 & \underline{89.14} & \textbf{1.33}\\
\rowcolor{ourback}\multirow{-4}{*}{\rotatebox[origin=c]{90}{\tiny\textbf{GLUE}}}& \method\ (ours) & \textbf{90.55} & 91.89/89.34 & \textbf{94.62} & \textbf{96.22} & \underline{71.02} & \textbf{88.81} & \textbf{90.44}/\textbf{93.10} & \underline{91.58}/\textbf{91.97} & \textbf{89.44} & \textbf{1.33}\\
\midrule
& Method & BoolQ {\tiny(acc)} & CB {\tiny(acc/F1)} & RTE {\tiny(acc)} & WiC {\tiny(acc)} & WSC {\tiny(acc)} & COPA {\tiny(acc)} & MultiRC {\tiny(F1a/EM)} & ReCoRD {\tiny(EM/F1)} & Avg & \#P {\tiny(M)}\\
\cmidrule(lr){2-12}
& LoRA & \textbf{83.43} & 85.71/76.13 & \textbf{85.92} & \underline{71.32} & \underline{77.88} & 87.00 & \textbf{75.24}/\textbf{38.30} & 83.64/84.39 & \underline{78.41} & \textbf{1.33}\\
\rowcolor{fullcol}& AdaLoRA ($\bar r{=}8$, repro) & \underline{82.94} & \underline{87.50}/\underline{78.08} & \textbf{85.92} & 70.06 & \textbf{78.85} & \textbf{94.00} & 60.19/14.80 & \underline{85.38}/\underline{85.93} & 77.21 & \textbf{1.33}\\
\rowcolor{ourback}\multirow{-4}{*}{\rotatebox[origin=c]{90}{\tiny\textbf{SuperGLUE}}}& \method\ (ours) & 82.81 & \textbf{89.29}/\textbf{80.37} & \textbf{85.92} & \textbf{71.79} & 75.96 & \underline{91.00} & \underline{71.46}/\underline{30.22} & \textbf{85.74}/\textbf{86.33} & \textbf{78.65} & \textbf{1.33}\\
\bottomrule
\end{tabular}}
\end{table}

\subsubsection{Structure Learning Stage: Extracting the Joint Structure}\label{sec:res-g1}
\paragraph{(1a) Multi-task all-at-once training.} Unlike the single-task runs of Table~\ref{tab:single_glue} (a separate adapter trained per task), here all methods are trained as a single multi-task adapter over the eight tasks at once. On GLUE (Table~\ref{tab:multi}) the baselines average $88.80$ and $88.42$; \method\ averages $89.07$, ahead of both. On SuperGLUE (Table~\ref{tab:multi}) \method\ averages $82.06$ (fair-rank at effective per-task rank $8$); HydraLoRA takes the overall average ($82.24$), driven mainly by its perfect score on the smallest task, yet \method\ reaches a comparable average without its MoE routing. Across both benchmarks the shared-Joint plus per-task Individual split adds usable capacity under joint training. We report \method's stronger rank-estimation configuration per benchmark here; the importance-based and permutation-test variants are compared in Appendix~\ref{sec:appmtl}.

\begin{table}[h]\centering\scriptsize
\caption{Multi-task results, all methods trained jointly on the eight tasks per benchmark. Number of
trainable parameters (\#P) are reported per single task head at sequence length $128$.}
\label{tab:multi}
\renewcommand{\arraystretch}{1.22}
\setlength{\aboverulesep}{0.05ex}
\setlength{\belowrulesep}{0.05ex}
\resizebox{\textwidth}{!}{%
\begin{tabular}{clcccccccccc}
\toprule
& Method & MNLI {\tiny(acc)} & QQP {\tiny(acc/F1)} & QNLI {\tiny(acc)} & SST-2 {\tiny(acc)} & CoLA {\tiny(Mcc)} & RTE {\tiny(acc)} & MRPC {\tiny(acc/F1)} & STS-B {\tiny(P/S)} & Avg & \#P {\tiny(M)}\\
\cmidrule(lr){2-12}
& HydraLoRA & 89.53 & 90.29/\textbf{87.33} & 94.14 & 95.41 & \underline{69.92} & \textbf{90.25} & 89.95/92.64 & 90.98/91.08 & \underline{88.80} & 2.90\\
& MTL-LoRA & \underline{89.93} & \textbf{90.36}/87.15 & \underline{94.49} & \underline{95.53} & 65.61 & \underline{89.17} & \textbf{90.93}/\textbf{93.36} & \underline{91.71}/\underline{91.68} & 88.42 & 2.66\\
\rowcolor{ourback}\multirow{-4}{*}{\rotatebox[origin=c]{90}{\tiny\textbf{GLUE}}}& \method\ (ours) & \textbf{90.10} & \underline{90.34}/\underline{87.32} & \textbf{94.58} & \textbf{95.99} & \textbf{70.23} & \underline{89.17} & \underline{90.20}/\underline{93.06} & \textbf{92.03}/\textbf{91.98} & \textbf{89.07} & \textbf{1.33}\\
\midrule
& Method & BoolQ {\tiny(acc)} & CB {\tiny(acc/F1)} & RTE {\tiny(acc)} & WiC {\tiny(acc)} & WSC {\tiny(acc)} & COPA {\tiny(acc)} & MultiRC {\tiny(F1a/EM)} & ReCoRD {\tiny(EM/F1)} & Avg & \#P {\tiny(M)}\\
\cmidrule(lr){2-12}
& HydraLoRA & \underline{85.54} & \textbf{100.00}/\textbf{100.00} & \textbf{88.45} & \underline{70.85} & 79.81 & \textbf{91.00} & \underline{76.94}/\underline{39.24} & 83.87/84.50 & \textbf{82.24} & 2.90\\
& MTL-LoRA & \textbf{85.66} & \underline{92.86}/89.24 & \underline{87.00} & \textbf{71.63} & \underline{80.77} & \underline{88.00} & 76.30/38.61 & \underline{86.27}/\underline{86.84} & 81.02 & 2.66\\
\rowcolor{ourback}\multirow{-4}{*}{\rotatebox[origin=c]{90}{\tiny\textbf{SuperGLUE}}}& \method\ (ours) & 84.86 & \underline{92.86}/\underline{89.32} & 85.92 & 69.28 & \textbf{87.50} & \textbf{91.00} & \textbf{77.28}/\textbf{42.39} & \textbf{86.76}/\textbf{87.28} & \underline{82.06} & \textbf{1.33}\\
\bottomrule
\end{tabular}}
\end{table}

\paragraph{(1b) Incremental multi-task adaptation.} We further show how Incremental \method\
(Algorithm~\ref{alg:incjive}) can be used as a \emph{continual-learning} procedure that mitigates
\emph{catastrophic forgetting}; a shared Joint is learned from a base cluster of tasks and each new task is
then folded in by refining both the shared Joint and the task's own Individual, so earlier tasks are
preserved rather than overwritten. \emph{Incremental folding needs only the stored low-rank Joint and
Individual factors of the earlier tasks together with the incoming task's data. Their training data is never
revisited, so \method\ applies even when the base-pool data no longer exists.} The full
study contrasts joint all-at-once training, incremental \method\ with single-task and multi-task
bases, and two sequential fine-tuning references (a single shared adapter and full backbone fine-tuning)
that keep no frozen shared structure and act as continual-learning lower bounds exposing forgetting, on
GLUE and SuperGLUE.
\emph{We recommend the incremental approach only when the task pool grows over time and the earlier tasks are
no longer available for joint all-at-once training, when the base-pool data is no longer accessible, or when
maintaining a separate head per task on the backbone for all-at-once training is too expensive; otherwise
all-at-once remains preferable.} We use Incremental \method\ as a remedy for such situations and defer this
study to Appendix~\ref{sec:appsinglebase} (Table~\ref{tab:incremental}) as an extension.

\subsubsection{Structure Reuse Stage}\label{sec:res-g2}
\begin{figure}[h]\centering
\includegraphics[width=\textwidth]{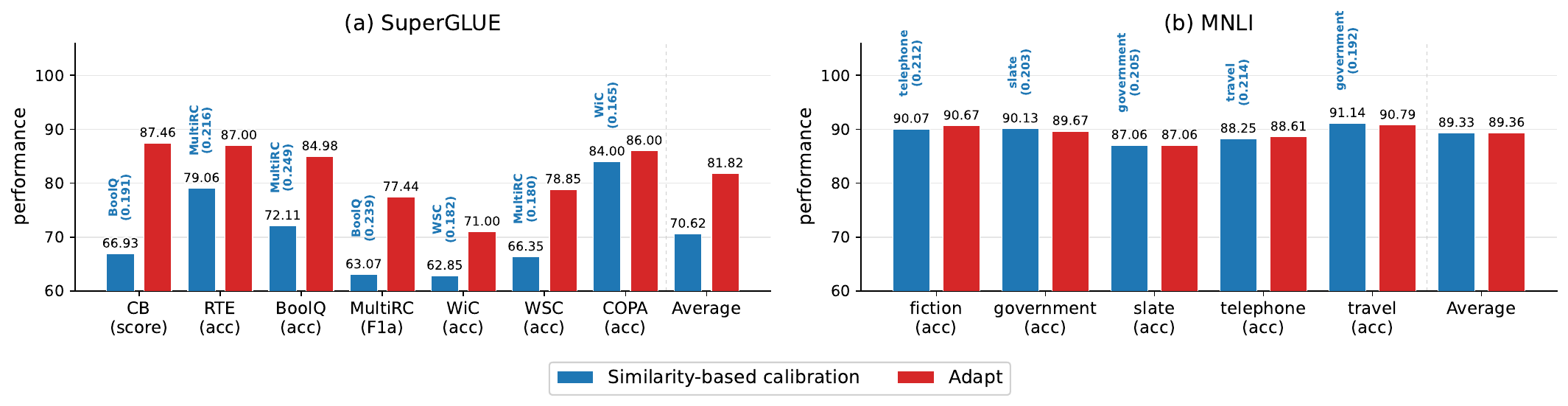}
\caption{Reuse on held-out tasks, our proposed similarity-based calibration vs adaptation.
In both pools, the similarity-based (blue) bar is annotated with the held-in donor it is routed to and the corresponding subspace overlap score.
(a) SuperGLUE cross-task pool; (b) MNLI cross-genre reuse.}
\label{fig:reuse}\label{fig:reuse_mnli}
\end{figure}

\paragraph{(2a) Similarity-based reuse.} We run a leave-one-out evaluation over the SuperGLUE pool, holding each task out in turn while \method\ is trained multi-task on the remaining six, freezing the per-layer Joint, and serving the held-out task through Alg.~\ref{alg:reuse}. Similarity-based reuse borrows a held-in task's Individual chosen purely by highest subspace overlap over the entire held-in cluster, with no task-type prior, re-scaling only its per-direction scale and training \emph{no} new Individual basis. We serve seven held-out tasks; routing follows subspace geometry rather than type labels, so the NLI tasks CB and RTE are routed to the boolean-QA donors BoolQ and MultiRC when those carry the highest overlap. The MNLI cross-genre pool is a natural fit here, since all five genres share the NLI type, so every held-out genre always has same-type donors and is routed to its highest-overlap genre (Figure~\ref{fig:reuse_mnli}). The pairwise subspace-overlap map shows the held-in Individuals are only weakly related to the held-out task, so no previously trained task is a close match. Even then, extracting the shared Joint and calibrating the closest Individual, adjusting only its scale and training no new basis, recovers most of the accuracy (Figure~\ref{fig:reuse}). Similarity-based reuse is therefore useful even when no near-identical task was trained before and when little or no data is available to fit a fresh Individual, since the reusable structure lives mainly in the shared Joint. A closer match still helps, since routing to a held-in task with higher subspace overlap yields higher performance; the strongest similarity-based reuse is on COPA, routed to WiC. Averaged over the pool, it falls short of adaptation on the heterogeneous SuperGLUE tasks ($70.62$ vs $81.82$) but matches it on the same-type MNLI genres ($89.33$ vs $89.36$). For the pairwise similarity scores refer to Appendix~\ref{sec:appexp}, Figure~\ref{fig:reuse_overlap}.

\paragraph{(2b) Adaptation reuse.} Adaptation keeps the frozen $\Jm^\star$ and fits a small new Individual for the held-out task from scratch, leaving the shared Joint untouched. It is the strong default and wins on every held-out SuperGLUE task, matching or approaching the all-eight-task multitask scores while adding only a lightweight per-task Individual; fitting a fresh Individual is thus the better choice even when a same-type twin is already available, except when the held-out task is very close to a held-in one, as with the MNLI genres, where similarity-based reuse matches adaptation while being cheaper. Unlike the incremental setting of (1b), which updates the shared Joint, here the Joint stays frozen and only the new Individual is trained, yet adaptation still recovers the multitask accuracy. Overlap stays low (near $0.2$) in both pools because each Individual is near-orthogonal to the shared Joint by construction and otherwise captures task-specific structure, \emph{so what governs reuse is not its absolute value but how much of the held-out task the shared Joint already covers; the same overlap level matches adaptation when the Joint suffices (the same-type MNLI genres, and COPA) yet lags when the task needs its own Individual (WiC, MultiRC)}.

\vspace{-1.2em}
\subsection{Rank Estimation and Interpretability}\label{sec:interp}
A budget-free permutation-rank estimator (Section~\ref{sec:permrank}) matches or improves on the importance allocator on multi-task SuperGLUE while removing the rank-budget hyperparameter; we report the full comparison in Appendix~\ref{sec:apprankcompare} (Table~\ref{tab:rank_est}). Because the split keeps the shared and task-specific parts as separate, orthogonal structures, each one can be measured, ablated, and visualized independently. We detail the interpretability hypotheses, metrics, and experiments in Appendix~\ref{sec:appexp}.

\section{Conclusion}\label{sec:conclusion}
We presented \method, a multi-task adapter whose every weight update is an additive low-rank
combination of a shared Joint and orthogonal per-task Individuals, with rank allocated adaptively
across the two pools.
Grounding this split in the linear JIVE decomposition, we recover it not by a least-squares fit but by
gradient-based nonlinear training of both parts as zero-initialized factors over a frozen backbone,
keeping the two parts interpretable and separately measurable. The decomposition supports a two-stage workflow: a Structure
Learning Stage that builds the Joint all at once or incrementally, and a Structure Reuse Stage that
freezes the Joint and serves a new task by reusing a related Individual with a cheap per-direction scale
or by training a small new one. On GLUE and SuperGLUE with DeBERTaV3-base, \method\ matches strong
single-task and multi-task low-rank baselines at a matched per-task effective rank and without any
mixture-of-experts routing, while remaining reusable and interpretable, and its incremental variant
mitigates catastrophic forgetting without revisiting earlier data.

\clearpage
\section*{Acknowledgements}
The authors would like to thank Can Goksen for insightful discussions and for his assistance with the experimental environment setup during the early stages of this work.

\section*{Use of AI Tools}
All ideas in this paper originate with the authors, are inspired by the authors' previous
work~\citep{abdali2021hjd}, and/or extend the related work cited herein. The core
implementation is entirely the authors' own work, including the \method\ adapter layer (\texttt{JIVELayer}),
the two-pool rank allocator, and the integration of both into DeBERTa. AI-based tools played only a
supporting role, limited to assisting with integrating the vendored baselines into our pipeline, refining the
phrasing of the authors' own text, assisting with \LaTeX\ drafting, polishing figures and tables, and
correcting grammar. Every technical contribution, claim, and conclusion was devised, checked, and approved
by the authors.

\section*{Ethics Statement}
The authors confirm that this work adheres to established ethical standards for machine learning
research. Our study is a
methodological contribution to parameter-efficient multi-task adaptation and is evaluated exclusively on
established public benchmarks (GLUE, SuperGLUE, and MNLI) under their standard licenses, so it involves no human
subjects, no private or personally identifiable data, and no new data collection or release. We foresee no
direct harmful application specific to the proposed adapter beyond the risks common to the underlying
pretrained language models, whose known limitations around bias and fairness carry over to any downstream
system; our method neither amplifies nor mitigates those properties by design. The authors declare no
conflicts of interest and no external sponsorship that could bias the reported results.

\section*{Reproducibility Statement}
We have made a deliberate effort to make every result reproducible. The method, including the additive
Joint and Individual decomposition and the orthogonality constraint, is specified in
Section~\ref{sec:method}, with the incremental and rank-allocation procedures given as explicit pseudocode
in Appendix~\ref{sec:appalg} alongside complete statements and proofs of the theoretical claims. The full
experimental setup, including datasets, task lists, optimizer, learning rate, budget, and the single fixed
random seed shared across all methods, is documented in Appendix~\ref{sec:appsetup}, and the extended
studies and their configurations appear in Appendix~\ref{sec:appext}. To further support exact
reproduction, we disclose the implementation of the core \texttt{JIVELayer}, the model-independent part that
is applicable to a variety of backbone models, together with its test units as anonymized supplementary
material, and defer the release of the full repository, including all training and
control scripts, to acceptance.

\bibliographystyle{iclr2027_conference}
\bibliography{references}

@inproceedings{abdali2021hjd,
  title     = {Semi-supervised Multi-aspect Detection of Misinformation Using Hierarchical Joint Decomposition},
  author    = {Abdali, Sara and Shah, Neil and Papalexakis, Evangelos E.},
  booktitle = {Machine Learning and Knowledge Discovery in Databases (ECML PKDD)},
  series    = {Lecture Notes in Computer Science},
  pages     = {406--422},
  publisher = {Springer},
  address   = {Cham},
  year      = {2021},
  doi       = {10.1007/978-3-030-67670-4_25}
}

@article{lock2013jive,
  title     = {Joint and Individual Variation Explained ({JIVE}) for Integrated Analysis of Multiple Data Types},
  author    = {Lock, Eric F. and Hoadley, Katherine A. and Marron, J. S. and Nobel, Andrew B.},
  journal   = {Annals of Applied Statistics},
  volume    = {7},
  number    = {1},
  pages     = {523--542},
  year      = {2013}
}

@inproceedings{zaken2022bitfit,
  title     = {{BitFit}: Simple Parameter-efficient Fine-tuning for Transformer-based Masked Language-models},
  author    = {Ben Zaken, Elad and Goldberg, Yoav and Ravfogel, Shauli},
  booktitle = {Annual Meeting of the Association for Computational Linguistics (ACL)},
  year      = {2022}
}

@inproceedings{zhang2023adalora,
  title     = {{AdaLoRA}: Adaptive Budget Allocation for Parameter-Efficient Fine-Tuning},
  author    = {Zhang, Qingru and Chen, Minshuo and Bukharin, Alexander and He, Pengcheng and Cheng, Yu and Chen, Weizhu and Zhao, Tuo},
  booktitle = {International Conference on Learning Representations (ICLR)},
  year      = {2023}
}

@inproceedings{hu2022lora,
  title     = {{LoRA}: Low-Rank Adaptation of Large Language Models},
  author    = {Hu, Edward J. and Shen, Yelong and Wallis, Phillip and Allen-Zhu, Zeyuan and Li, Yuanzhi and Wang, Shean and Wang, Lu and Chen, Weizhu},
  booktitle = {International Conference on Learning Representations (ICLR)},
  year      = {2022}
}

@inproceedings{wang2019glue,
  title     = {{GLUE}: A Multi-Task Benchmark and Analysis Platform for Natural Language Understanding},
  author    = {Wang, Alex and Singh, Amanpreet and Michael, Julian and Hill, Felix and Levy, Omer and Bowman, Samuel R.},
  booktitle = {International Conference on Learning Representations (ICLR)},
  year      = {2019}
}

@inproceedings{wang2019superglue,
  title     = {{SuperGLUE}: A Stickier Benchmark for General-Purpose Language Understanding Systems},
  author    = {Wang, Alex and Pruksachatkun, Yada and Nangia, Nikita and Singh, Amanpreet and Michael, Julian and Hill, Felix and Levy, Omer and Bowman, Samuel R.},
  booktitle = {Advances in Neural Information Processing Systems (NeurIPS)},
  year      = {2019}
}

@inproceedings{houlsby2019adapter,
  title     = {Parameter-Efficient Transfer Learning for {NLP}},
  author    = {Houlsby, Neil and Giurgiu, Andrei and Jastrzebski, Stanislaw and Morrone, Bruna and de Laroussilhe, Quentin and Gesmundo, Andrea and Attariyan, Mona and Gelly, Sylvain},
  booktitle = {International Conference on Machine Learning (ICML)},
  year      = {2019}
}

@inproceedings{pfeiffer2021adapterfusion,
  title     = {{AdapterFusion}: Non-Destructive Task Composition for Transfer Learning},
  author    = {Pfeiffer, Jonas and Kamath, Aishwarya and R{\"u}ckl{\'e}, Andreas and Cho, Kyunghyun and Gurevych, Iryna},
  booktitle = {Conference of the European Chapter of the Association for Computational Linguistics (EACL)},
  year      = {2021}
}

@inproceedings{mahabadi2021hyperformer,
  title     = {Parameter-Efficient Multi-Task Fine-Tuning for Transformers via Shared Hypernetworks},
  author    = {Karimi Mahabadi, Rabeeh and Ruder, Sebastian and Dehghani, Mostafa and Henderson, James},
  booktitle = {Annual Meeting of the Association for Computational Linguistics (ACL)},
  year      = {2021}
}

@article{ponti2023polytropon,
  title     = {Combining Modular Skills in Multitask Learning},
  author    = {Ponti, Edoardo M. and Sordoni, Alessandro and Bengio, Yoshua and Reddy, Siva},
  journal   = {arXiv preprint arXiv:2202.13914},
  year      = {2022}
}

@inproceedings{yang2024mtllora,
  title     = {{MTL-LoRA}: Low-Rank Adaptation for Multi-Task Learning},
  author    = {Yang, Yaming and Muhtar, Dilxat and Shen, Yelong and Zhan, Yuefeng and Liu, Jianfeng and Wang, Yujing and Sun, Hao and Deng, Denvy and Sun, Feng and Zhang, Qi and Chen, Weizhu and Tong, Yunhai},
  booktitle = {Proceedings of the AAAI Conference on Artificial Intelligence (AAAI)},
  year      = {2025}
}

@inproceedings{tian2024hydralora,
  title     = {{HydraLoRA}: An Asymmetric {LoRA} Architecture for Efficient Fine-Tuning},
  author    = {Tian, Chunlin and Shi, Zhan and Guo, Zhijiang and Li, Li and Xu, Chengzhong},
  booktitle = {Advances in Neural Information Processing Systems (NeurIPS)},
  year      = {2024}
}

@inproceedings{ning2025mode,
  title     = {{MoDE}: Effective Multi-Task Parameter Efficient Fine-Tuning with a Mixture of Dyadic Experts},
  author    = {Ning, Lin and Lara, Harsh and Guo, Meiqi and Rastogi, Abhinav},
  booktitle = {Findings of the Association for Computational Linguistics: NAACL},
  year      = {2025}
}

@inproceedings{achille2019task2vec,
  title     = {{Task2Vec}: Task Embedding for Meta-Learning},
  author    = {Achille, Alessandro and Lam, Michael and Tewari, Rahul and Ravichandran, Avinash and Maji, Subhransu and Fowlkes, Charless C. and Soatto, Stefano and Perona, Pietro},
  booktitle = {IEEE/CVF International Conference on Computer Vision (ICCV)},
  year      = {2019}
}

@inproceedings{saha2021gpm,
  title     = {Gradient Projection Memory for Continual Learning},
  author    = {Saha, Gobinda and Garg, Isha and Roy, Kaushik},
  booktitle = {International Conference on Learning Representations (ICLR)},
  year      = {2021}
}

@article{bjorck1973angles,
  title     = {Numerical Methods for Computing Angles Between Linear Subspaces},
  author    = {Bj{\"o}rck, {\AA}ke and Golub, Gene H.},
  journal   = {Mathematics of Computation},
  volume    = {27},
  number    = {123},
  pages     = {579--594},
  year      = {1973}
}

@inproceedings{he2023debertav3,
  title     = {{DeBERTaV3}: Improving {DeBERTa} using {ELECTRA}-Style Pre-Training with Gradient-Disentangled Embedding Sharing},
  author    = {He, Pengcheng and Gao, Jianfeng and Chen, Weizhu},
  booktitle = {International Conference on Learning Representations (ICLR)},
  year      = {2023}
}

@inproceedings{aghajanyan2021intrinsic,
  title     = {Intrinsic Dimensionality Explains the Effectiveness of Language Model Fine-Tuning},
  author    = {Aghajanyan, Armen and Gupta, Sonal and Zettlemoyer, Luke},
  booktitle = {Annual Meeting of the Association for Computational Linguistics (ACL)},
  pages     = {7319--7328},
  year      = {2021}
}

@inproceedings{li2021prefix,
  title     = {Prefix-Tuning: Optimizing Continuous Prompts for Generation},
  author    = {Li, Xiang Lisa and Liang, Percy},
  booktitle = {Annual Meeting of the Association for Computational Linguistics (ACL)},
  pages     = {4582--4597},
  year      = {2021}
}

@inproceedings{lester2021prompt,
  title     = {The Power of Scale for Parameter-Efficient Prompt Tuning},
  author    = {Lester, Brian and Al-Rfou, Rami and Constant, Noah},
  booktitle = {Conference on Empirical Methods in Natural Language Processing (EMNLP)},
  pages     = {3045--3059},
  year      = {2021}
}

@inproceedings{williams2018mnli,
  title     = {A Broad-Coverage Challenge Corpus for Sentence Understanding through Inference},
  author    = {Williams, Adina and Nangia, Nikita and Bowman, Samuel R.},
  booktitle = {North American Chapter of the Association for Computational Linguistics (NAACL)},
  year      = {2018}
}

@inproceedings{liu2024dora,
  title     = {{DoRA}: Weight-Decomposed Low-Rank Adaptation},
  author    = {Liu, Shih-Yang and Wang, Chien-Yi and Yin, Hongxu and Molchanov, Pavlo and Wang, Yu-Chiang Frank and Cheng, Kwang-Ting and Chen, Min-Hung},
  booktitle = {International Conference on Machine Learning (ICML)},
  year      = {2024}
}

@inproceedings{wang2023olora,
  title     = {Orthogonal Subspace Learning for Language Model Continual Learning},
  author    = {Wang, Xiao and Chen, Tianze and Ge, Qiming and Xia, Han and Bao, Rong and Zheng, Rui and Zhang, Qi and Gui, Tao and Huang, Xuanjing},
  booktitle = {Findings of the Association for Computational Linguistics: EMNLP},
  year      = {2023}
}

@inproceedings{huang2024lorahub,
  title     = {{LoraHub}: Efficient Cross-Task Generalization via Dynamic {LoRA} Composition},
  author    = {Huang, Chengsong and Liu, Qian and Lin, Bill Yuchen and Pang, Tianyu and Du, Chao and Lin, Min},
  booktitle = {Conference on Language Modeling (COLM)},
  year      = {2024}
}

@article{feng2018ajive,
  title     = {Angle-Based Joint and Individual Variation Explained},
  author    = {Feng, Qing and Jiang, Meilei and Hannig, Jan and Marron, J. S.},
  journal   = {Journal of Multivariate Analysis},
  volume    = {166},
  pages     = {241--265},
  year      = {2018}
}

\clearpage
\appendix

\section{Extended Related Work}\label{sec:apprelated}
Beyond the counterparts compared in the main text, several other parameter-efficient and
multi-task adaptation families are related to \method; Table~\ref{tab:related} summarizes how they
compare along the axes most relevant to our contribution.

Prompt- and prefix-based tuning prepend learned continuous vectors while freezing the
backbone~\citep{li2021prefix,lester2021prompt}. BitFit~\citep{zaken2022bitfit} updates only the bias
terms. DoRA~\citep{liu2024dora} decomposes each pretrained
weight into a magnitude and a direction and applies the low-rank update to the direction alone.
Adapters~\citep{houlsby2019adapter} insert per-task bottleneck layers. These methods target a single
task at a time and, for the adapter-insertion variants, alter the architecture by adding modules to
every layer, which is not always feasible in deployment.

\paragraph{Multi-task adaptation.}
AdapterFusion~\citep{pfeiffer2021adapterfusion} fuses per-task bottleneck layers,
HyperFormer(++)~\citep{mahabadi2021hyperformer} generates per-task adapters from a shared
meta-network, Polytropon~\citep{ponti2023polytropon} routes over an inventory of LoRA adapters, and
MoDE~\citep{ning2025mode} mixes shared and task experts. Closer to our orthogonality and reuse
goals, O-LoRA~\citep{wang2023olora} learns each new task in a low-rank subspace kept orthogonal to
those of earlier tasks to resist forgetting in continual learning, and
LoRAHub~\citep{huang2024lorahub} composes separately trained LoRA modules to generalize to unseen
tasks.

\begin{table}[htbp]\centering
\caption{Comparison of \method\ with representative adaptation methods.}
\label{tab:related}
\resizebox{\textwidth}{!}{%
\begin{tabular}{lcccccc}
\toprule
Method & Multi-task & Shared+specific & Orthogonal & Adaptive rank & Backbone-preserving & Reusable\\
\midrule
Prefix/Prompt~\citep{li2021prefix,lester2021prompt} &            &            &            &            &            &            \\
LoRA~\citep{hu2022lora}                              &            &            &            &            & \checkmark &            \\
AdaLoRA~\citep{zhang2023adalora}                     &            &            &            & \checkmark & \checkmark &            \\
DoRA~\citep{liu2024dora}                             &            &            &            &            & \checkmark &            \\
Adapters~\citep{houlsby2019adapter}                  &            &            &            &            &            &            \\
AdapterFusion~\citep{pfeiffer2021adapterfusion}      & \checkmark & \checkmark &            &            &            & \checkmark \\
HyperFormer(++)~\citep{mahabadi2021hyperformer}      & \checkmark & \checkmark &            &            &            &            \\
Polytropon~\citep{ponti2023polytropon}              & \checkmark & \checkmark &            &            & \checkmark & \checkmark \\
MTL-LoRA~\citep{yang2024mtllora}                     & \checkmark & \checkmark &            &            & \checkmark &            \\
HydraLoRA~\citep{tian2024hydralora}                  & \checkmark & \checkmark &            &            & \checkmark &            \\
MoDE~\citep{ning2025mode}                            & \checkmark & \checkmark &            &            & \checkmark &            \\
O-LoRA~\citep{wang2023olora}                         & \checkmark &            & \checkmark &            & \checkmark &            \\
LoRAHub~\citep{huang2024lorahub}                     & \checkmark &            &            &            & \checkmark & \checkmark \\
\textbf{\method} (ours)            & \checkmark & \checkmark & \checkmark & \checkmark & \checkmark & \checkmark \\
\bottomrule
\end{tabular}}
\end{table}

\section{Algorithms and Proofs}\label{sec:appalg}
This appendix presents the algorithms underlying \method.

\subsection{Rank Allocation Algorithms}\label{sec:apprankalgs}
\method\ sets the per-pool ranks in one of two ways: the importance-based allocator
and the budget-free permutation-test estimator of
linear JIVE~\citep{lock2013jive} (Section~\ref{sec:permrank}). This appendix details both (the importance-based allocator is summarized in Section~\ref{sec:alloc}).

\subsubsection{Adaptive Rank Allocation Algorithm}\label{sec:apprank}
We adopt AdaLoRA's importance-based
rank allocation~\citep{zhang2023adalora} to the multi-task, per-structure setting. The shared Joint
of each layer and each task's Individual are independent \emph{structures}; every structure
owns a vector of singular values $E$. This subsection gives the importance scoring, the per-pool
budget schedule, and the full allocation algorithm.

\paragraph{Sensitivity and importance.}
Per factor parameter $p$ we track a smoothed sensitivity and uncertainty:
\begin{align}
  I(p) &= |p\odot\nabla p|, &
  \overline{I}&\leftarrow \beta_1\overline{I}+(1-\beta_1)I(p), &
  \overline{U}&\leftarrow \beta_2\overline{U}+(1-\beta_2)\bigl|I(p)-\overline{I}\bigr|,
\end{align}
and score $s(p)=\overline{I}\odot\overline{U}$. The score of singular value $i$ of a
structure combines its $E$ entry with the mean scores of the corresponding $V$ row and $U$
column:
\begin{equation}
  s_i \;=\; s(E)_i \;+\; \mathrm{mean}_{\text{in}} s(V)_{i,:} \;+\; \mathrm{mean}_{\text{out}} s(U)_{:,i}.
\end{equation}

\paragraph{Per-structure budgeting.}
The Joint pool and the Individual pool are budgeted \emph{separately} so many per-task
structures cannot starve the single shared structure. Each pool's active-rank budget
decays via the \textsc{Triplet} schedule from its full size to its target between the initial and final warmups
(Alg.~\ref{alg:alloc}); within a pool, the lowest-scoring singular values across all its
structures are zeroed to meet the budget. Optionally an explicit total budget is split
proportionally between the pools.

\begin{algorithm}[h]
\caption{\textsc{Triplet}$(N,\tau,g)$: per-pool active-rank budget schedule}\label{alg:cubic}
\fontsize{6.2}{7.3}\selectfont
\begin{algorithmic}[1]
\Require pool full size $N$, pool target $\tau$, step $g$; warmups $t_i,t_f$; total steps $T_{\text{total}}$
\Ensure active-rank budget $b\in[\tau,N]$
\If{$g\le t_i$} \Return $N$ \Comment{initial warmup: keep full rank}
\ElsIf{$g\ge T_{\text{total}}-t_f$} \Return $\tau$ \Comment{final warmup: hold at target}
\Else
  \State $p\gets(g-t_i)\,/\,(T_{\text{total}}-t_f-t_i)$ \Comment{progress $\in(0,1)$}
  \State \Return $\bigl\lfloor\,\tau+(N-\tau)(1-p)^3\,\bigr\rfloor$ \Comment{triplet decay $N\!\to\!\tau$}
\EndIf
\end{algorithmic}
\end{algorithm}

\begin{algorithm}[h]
\caption{Adaptive rank allocation step (allocator update-and-mask)}\label{alg:alloc}
\fontsize{6.2}{7.3}\selectfont
\begin{algorithmic}[1]
\Require model, global step $g$; totals $N_J,N_I$; targets $\tau_J,\tau_I$; warmups $t_i,t_f$; interval $\Delta$
\State \iphase{importance update} if $g < T_{\text{total}}-t_f$, update EMAs $\overline I,\overline U$ for all Joint/Individual factors
\State \iphase{budget schedule} $b_J\gets\textsc{Triplet}(N_J,\tau_J,g)$;\quad $b_I\gets\textsc{Triplet}(N_I,\tau_I,g)$ \Comment{Alg.~\ref{alg:cubic}, per pool}
\State $\text{mask}\gets(g>t_i)\wedge\bigl(g>T_{\text{total}}-t_f\ \vee\ g\bmod\Delta=0\bigr)$;\quad if $\neg\text{mask}$ \Return
\State \iphase{score and threshold} pool$_J\gets$ concat of $\{s_i\}$ over all Joint structures; pool$_I\gets$ same over all Individual structures
\State $\theta_J\gets$ $(N_J-b_J)$-th smallest of pool$_J$;\quad $\theta_I\gets$ $(N_I-b_I)$-th smallest of pool$_I$
\State \iphase{mask singular values} for each Joint structure, zero $E_J[i]$ where $s_i\le\theta_J$; for each Individual $(\text{layer},t)$, zero $E_{I,t}[i]$ where $s_i\le\theta_I$
\State record the kept ranks per structure
\end{algorithmic}
\end{algorithm}

\noindent\textbf{Consistent scoring.} Every singular value is ranked by the same importance score $s_i$ above, identically for the Joint and Individual~pools.

\subsubsection{Permutation-test Rank Estimation Algorithm}\label{sec:apppermrank}
This is the budget-free alternative from Section~\ref{sec:permrank}. It adopts the permutation
test of linear JIVE~\citep{lock2013jive} for weight space and runs it \emph{once, before training},
as a one-shot estimator that fixes the ranks up front. For each task $t$ (and for the shared module) we average the gradient of that
task's loss with respect to the frozen base weight over a few batches, giving one weight-shaped
block $X_t$ per task. Passing the blocks $\{X_t\}$ to Alg.~\ref{alg:permrank} yields a Joint rank
and a per-task Individual rank. We then reduce these per-module estimates to the two ranks the
adapter exposes. The Joint rank is the median across modules of each module's Joint rank. The
Individual rank is computed in two steps. Within each module we average the per-task Individual
ranks (the per-module mean), then take the median of those per-module means across modules. The
adapter is rebuilt at these two ranks, and nothing is pruned during training.

\begin{algorithm}[h]
\caption{Permutation-test rank estimation~\citep{lock2013jive}}\label{alg:permrank}
\fontsize{6.2}{7.3}\selectfont
\begin{algorithmic}[1]
\Require per-task blocks $\{X_t\}_{t=1}^{k}$ (each the module's averaged frozen-weight gradient, $d\times n$, shared columns); permutations $B$; level $\alpha$; max outer iters $M$
\Ensure Joint rank $r_J$; per-task Individual ranks $\{r_{I,t}\}$
\State $r_J\gets 0$;\ \ $r_{I,t}\gets 0\ \forall t$;\ \ $J_t\gets 0,\ A_t\gets 0$ \Comment{initialize estimates}
\For{$m=1,\dots,M$}
  \State $S\gets [\,X_1-A_1;\ \dots;\ X_k-A_k\,]$ \Comment{stack the individual-removed blocks}
  \For{$b=1,\dots,B$} permute each block's columns independently, restack, record $\sigma^{(b)}_{\max}$ \EndFor
  \State $q_J\gets$ the $(1-\alpha)$ quantile of $\{\sigma^{(b)}_{\max}\}$
  \State $r_J\gets \max\big(r_J,\ \#\{\text{leading singular values of }S > q_J\}\big)$ \Comment{count from the top; non-decreasing}
  \For{$t=1,\dots,k$}
    \State $q_t\gets (1-\alpha)$ quantile of top singular values of $X_t-J_t$ with entries permuted within each row
    \State $r_{I,t}\gets \#\{\text{leading singular values of }X_t-J_t > q_t\}$ \Comment{count from the top}
  \EndFor
  \State $\{J_t\},\{A_t\}\gets\textsc{JIVEIter}(\{X_t\},\,r_J,\,\{r_{I,t}\})$\footnotemark \Comment{alternating low-rank fit, $A_t\perp J$}
  \State \textbf{break if} $r_J$ and all $r_{I,t}$ are unchanged
\EndFor
\State \Return $r_J,\ \{r_{I,t}\}$ \Comment{aggregate across modules to the adapter's $r_J,r_I$}
\end{algorithmic}
\end{algorithm}
\footnotetext{\textsc{JIVEIter} is the classic JIVE alternating decomposition of \citet{lock2013jive}.}

\subsection{Principal-Angle Subspace Overlap}\label{sec:appoverlap}
This appendix justifies the overlap score for nearest-task routing in
Structure Reuse Stage (2a) {\color{modecol}\textsc{similarity}} (Eq.~\eqref{eq:overlap}). The singular values of $G^\top Q$ are exactly the cosines
of the principal angles between the two subspaces~\citep{bjorck1973angles}.

\paragraph{Setup.} Let $G\in\R^{n\times r}$ and $Q\in\R^{n\times r}$ have orthonormal columns,
$G^\top G=Q^\top Q=\mathbb{I}_r$, spanning subspaces $\mathcal G=\mathrm{col}(G)$ and
$\mathcal Q=\mathrm{col}(Q)$. The principal angles
$0\le\theta_1\le\cdots\le\theta_r\le\pi/2$ between $\mathcal G$ and $\mathcal Q$ are defined
recursively by
\begin{equation}
  \cos\theta_i=\max_{\substack{u\in\mathcal G,\ v\in\mathcal Q\\ \|u\|=\|v\|=1}}\ u^\top v
  \quad\text{s.t.}\quad u\perp u_j,\ v\perp v_j\ \ (j<i),
  \label{eq:pa-def}
\end{equation}
with maximizers $u_i,v_i$ the $i$-th principal directions.

\paragraph{Proposition.} Let $M=G^\top Q\in\R^{r\times r}$ have SVD $M=A\,\diag(\sigma_1,\dots,\sigma_r)\,B^\top$
with $\sigma_1\ge\cdots\ge\sigma_r\ge0$. Then $\sigma_i=\cos\theta_i$ for all $i$, and the
principal directions are $u_i=Ga_i$, $v_i=Qb_i$ (the $a_i,b_i$ being the columns of $A,B$).
Consequently
\begin{equation}
  \|G^\top Q\|_F^2=\sum_{i=1}^{r}\sigma_i^2=\sum_{i=1}^{r}\cos^2\theta_i,
  \label{eq:overlap-proof}
\end{equation}
which is the (unnormalized) overlap in Eq.~\eqref{eq:overlap}. This is the classical
characterization of principal angles between subspaces~\citep{bjorck1973angles}; we include a short
proof for completeness.

\paragraph{Proof.} For unit vectors $u=Ga\in\mathcal G$ and $v=Qb\in\mathcal Q$, orthonormality
of $G,Q$ gives $\|u\|=\|a\|$ and $\|v\|=\|b\|$, so $u,v$ are unit iff $a,b$ are unit. Their inner
product is
\[
  u^\top v=(Ga)^\top(Qb)=a^\top(G^\top Q)b=a^\top M b .
\]
Thus maximizing $u^\top v$ over unit $u\in\mathcal G,\ v\in\mathcal Q$ is equivalent to maximizing
$a^\top M b$ over unit $a,b\in\R^{r}$, whose optimum is the largest singular value $\sigma_1$ of
$M$, attained at the leading singular vectors $a_1,b_1$. The orthogonality constraints in
Eq.~\eqref{eq:pa-def} ($u\perp u_j\Leftrightarrow a\perp a_j$ and $v\perp v_j\Leftrightarrow
b\perp b_j$, again by orthonormality of $G,Q$) restrict the $i$-th step to the orthogonal
complement of the previous singular vectors, whose optimum is the next singular value
$\sigma_i$, attained at $a_i,b_i$. Hence $\cos\theta_i=\sigma_i$ and $u_i=Ga_i,\ v_i=Qb_i$.
Finally $\|M\|_F^2=\sum_i\sigma_i^2=\sum_i\cos^2\theta_i$, giving Eq.~\eqref{eq:overlap-proof}.
\hfill$\square$

\paragraph{Remark on the range of the score.} Since $G,Q$ have orthonormal columns, $\|Gx\|=\|x\|$,
so $M=G^\top Q$ has operator norm $\|M\|_2\le1$ and every $\sigma_i\in[0,1]$; thus each
$\theta_i=\arccos\sigma_i\in[0,\pi/2]$ is a well-defined angle. The normalized score
$\tfrac1r\sum_i\cos^2\theta_i$ therefore lies in $[0,1]$, equals $1$ iff $\mathcal G=\mathcal Q$
(all $\theta_i=0$), and equals $0$ iff the subspaces are orthogonal (all $\theta_i=\pi/2$).

\section{Experimental Setup}\label{sec:appsetup}
This appendix documents the experimental setup and configuration.
Unless otherwise stated, every result reported in the paper uses one fixed random seed, shared by
\method\ and all baselines, so that every method sees the same seed, data order, and budget; the reported
differences are therefore controlled comparisons that are not confounded by seed variation, and fixing the
seed also makes each run exactly reproducible.
Table~\ref{tab:baselines} compares the capacity structure of \method\ and the baselines we run.

\begin{table}[h]\centering\scriptsize
\caption{Capacity structure and single-task cost ($T{=}1$) of \method\ and the baselines.}
\label{tab:baselines}
\resizebox{0.72\textwidth}{!}{%
\begin{tabular}{lllcc}
\toprule
Method & Shared cap. & Task-specific cap. & Adaptive rank & \#P {\tiny(M)}\\
\midrule
LoRA         & none          & $BA$       & no (fixed $r$) & 1.33\\
AdaLoRA      & none          & $U\diag(E)V$           & yes (cross-layer) & 1.33\\
HydraLoRA    & shared $A$    & per-task heads $B_t$   & no (fixed $r$) & 2.90\\
MTL-LoRA     & shared $A$    & per-task $\Lambda_t$ $+$ up-proj & no (fixed $r$) & 2.66\\
\textbf{\method} & Joint $\Jm$ & Individual $\Im_t$ ($\perp\Jm$) & yes & 1.33\\
\textbf{\method}\ similarity reuse & frozen $\Jm$ & routed $\Im_{t'}$ & no & 0.00\\
\textbf{\method}\ adapt reuse & frozen $\Jm$ & fresh $\Im_{\hat t}$ & yes & 0.66\\
\bottomrule
\end{tabular}}
\end{table}

\subsection{Compute and Software Environment}\label{sec:appcompute}
All training and experiments run on a single Azure Machine Learning node of type
\texttt{Standard\_NC24ads\_A100\_v4}, which provides one NVIDIA A100 PCIe GPU with $80$\,GB of
on-board memory, $24$ AMD EPYC (Milan) vCPUs, and $220$\,GiB of host RAM. Every job uses a single
GPU; no multi-GPU or multi-node parallelism is required, since the DeBERTaV3-base backbone with
low-rank adapters has a small footprint and trains comfortably within the $80$\,GB budget, with ample
headroom for the multi-task loader that keeps all task heads resident at once. The software stack is
the Azure ML curated ACPT image built on PyTorch $2.2$ and CUDA $12.1$, on top of which we install a
patched HuggingFace Transformers in editable mode. \method\ and every baseline share this identical
image, hardware, and driver stack, so reported differences reflect the method rather than the
environment.

\subsection{Parameters and Invariants}\label{sec:impl}
\paragraph{Parameters and invariants.}
\textbf{Per-layer hyperparameters.} The ranks $r_I,r_J$, scalings $\alpha_I,\alpha_J$, dropout, and
task count $T$. \textbf{Trainable parameters.} Shared Joint $V_J,E_J,U_J$ and per-task Individual
$V_{I,t},E_{I,t},U_{I,t}$ over $T$ tasks; the base weight $\Wm$ and per-pool rank counts stay frozen.
\textbf{Verified invariants.} $\Delta\Wm_t=0$ at init; the routed forward equals per-sample
(Joint\,$+$\,its Individual) to $\sim\!10^{-6}$; embedding routing matches; Joint gets gradient from
all tasks, Individual $t$ only from task-$t$ samples; single-task merge/unmerge is exact; the allocator
prunes each pool to target and the forward still runs. \textbf{Homogeneous micro-batch variant.} When
the loader cannot mix tasks, accumulating one per-task micro-batch each before a single step recovers
the same gradient, with Joint seeing the summed gradient and each Individual only its own.

\subsection{Configuration and Rank Budgets}\label{sec:appmtl}
This subsection documents the configuration used for the multi-task table
(Table~\ref{tab:multi}). All methods share the same backbone, injected modules
(query, key, value, intermediate, layer.output, attention.output), and optimization
schedule; they differ only in how adapter rank is organized.

\paragraph{Matched per-task effective rank.} The controls and \method\ are matched on the
\emph{per-task effective rank}, i.e.\ the adapter rank any single task sees at inference, not on
the nominal target-rank argument. AdaLoRA uses a single shared SVD pool with a target rank of
$8$ (its allocator's final average rank per structure). \method\ multitask
splits capacity across per-structure rank pools, a shared Joint and a per-task Individual, each with
a target rank of $4$, so every task still sees Joint($4$)${+}$Individual($4$)${=}8$ active
rank. LoRA uses a single fully-shared adapter at the same effective rank. All three start
from an initial rank of $12$ per structure (both the Joint and Individual pools of \method\ start at
$12$, as do the LoRA and AdaLoRA adapters) and scaling
$\alpha{=}16$.

\paragraph{Rank budget selection for fair comparison.} We match the per-task effective rank
across methods rather than the raw parameter count, which differs by structure. For a target average rank $\bar r$, AdaLoRA's budget is
$b=\bar r\cdot n$ over $n$ injected matrices; \method\ splits this between the Joint pool ($r_J$)
and the Individual pool ($\sum_t r_{I,t}$) so that $r_J+\tfrac1T\sum_t r_{I,t}\approx\bar r$,
keeping per-task capacity comparable. We vary $\bar r\in\{2,4,8\}$ (final total ranks
$144,288,576$ over 72 matrices) with initial ranks $\{4,6,12\}$, matching~AdaLoRA.
Table~\ref{tab:mtl_rank_budget} details the matched rank budget and shared optimization schedule.

\begin{table}[h]\centering\small
\caption{Multi-task run configuration at matched per-task effective rank $8$, with AdaLoRA at one
shared rank-$8$ pool and \method\ at a shared Joint plus per-task Individual of rank $4$ each.}
\label{tab:mtl_rank_budget}
\resizebox{0.72\columnwidth}{!}{%
\begin{tabular}{llll}
\toprule
\multicolumn{4}{c}{\textbf{Rank budget (matched per-task effective rank $8$)}}\\
\midrule
Method & Pools & Init rank & \texttt{target\_rank}\\
LoRA (fully shared)    & 1 shared                  & 12 & (fixed)\\
AdaLoRA (fully shared) & 1 shared                  & 12 & 8\\
\method\ (Importance-based)     & Joint $+$ $T$ Individuals & 12 & 4 (per structure, $4{+}4{=}8$)\\
\midrule
\multicolumn{4}{c}{\textbf{Shared optimization schedule (all runs; allocator shared by AdaLoRA and \method)}}\\
\midrule
Learning rate & $5\times10^{-4}$ & Seed & $6$\\
Weight decay & $0$ & Classifier dropout & $0.1$\\
Batch size & $8$ & Orthogonality reg.\ coef. & $0.1$\\
Max sequence length & $256$ & Allocator init warmup & $4000$\\
Training steps & $20{,}000$ & Allocator mask interval & $100$\\
Warmup steps & $1{,}000$ & Allocator final warmup & $8000$\\
Eval interval (steps) & $2{,}000$ & & \\
\midrule
\multicolumn{4}{c}{\textbf{Method-specific settings (structured baselines)}}\\
\midrule
Setting & MTL-LoRA & HydraLoRA & \\
Rank & $8$ & $8$ & \\
Scaling $\alpha$ & $16$ & $16$ & \\
Dropout & $0.1$ & $0.1$ & \\
B matrices / experts & $3$ & $3$ & \\
B scale & $1.0$ & n/a & \\
\bottomrule
\end{tabular}%
}
\end{table}
\vspace{-1.2em}

\paragraph{Per-sample routing in DeBERTa's disentangled attention.} DeBERTaV3's disentangled
attention adds position-based query/key projections for the relative-position embeddings on top of
the usual content query/key/value projections. \method\ multitask routes each sample to its
Individual, and this routing must reach the position projections, not only the content ones.
Leaving the position projections unrouted makes the multi-task model fail to converge; routing every
attention projection fixes this and yields the reported GLUE average ($89.07$).

\section{Extended Experiments}\label{sec:appext}
\subsection{Incremental \method\ as continual learning}\label{sec:appsinglebase}
Table~\ref{tab:incremental} reports the full incremental study, comparing joint all-at-once \method,
incremental \method\ (Algorithm~\ref{alg:incjive}), which learns the shared Joint from a base cluster and then
folds each remaining task in by refining both the shared Joint and its own Individual, and two sequential
fine-tuning references (a single shared adapter and full backbone fine-tuning) that keep no frozen shared
structure and act as continual-learning lower bounds. To examine the effect of the base cluster size, we form
the initial Joint from a cluster of size one (single-task base) or larger (multi-task base); for GLUE these
are \{MNLI\} and \{MNLI, QQP, SST-2, QNLI\}, with the remaining tasks streamed in both cases. In practice the
cluster size is dictated by model capacity, such as how many task heads the backbone can host for
all-at-once training, and by whether the base-pool data is still available. All incremental and sequential
runs use a matched rank budget with the same allocator and hyperparameters as the
main experiments, scored on all tasks after the full stream. Both incremental variants score below
joint all-at-once training, more so on GLUE ($81.13$ and $79.54$ vs $89.07$) than on SuperGLUE ($80.72$ and
$73.69$ vs $82.06$), yet they far outperform the two sequential references (GLUE $61.11$ and $69.55$;
SuperGLUE $68.36$ and $65.60$), whose shared adapter or fully fine-tuned backbone is overwritten by later
tasks; \method\ avoids this \emph{continual-learning} failure by keeping the shared Joint and refreshing each
orthogonal Individual, so earlier tasks are preserved rather than forgotten. \emph{The base cluster size
matters more on the heterogeneous SuperGLUE tasks, where the multi-task base ($80.72$) clearly leads the
single-task base ($73.69$) than on GLUE, where the single-task base ($79.54$) already
recovers almost all of
the multi-task base accuracy ($81.13$).}

\begin{table}[H]\centering\scriptsize
\caption{Incremental \method\ as a continual-learning procedure on GLUE and SuperGLUE, using a matched rank
budget.}
\label{tab:incremental}
\renewcommand{\arraystretch}{1.15}
\resizebox{\textwidth}{!}{%
\begin{tabular}{clccccccccc}
\toprule
& Method & MNLI {\tiny(acc)} & QQP {\tiny(acc/F1)} & QNLI {\tiny(acc)} & SST-2 {\tiny(acc)} & CoLA {\tiny(Mcc)} & RTE {\tiny(acc)} & MRPC {\tiny(acc/F1)} & STS-B {\tiny(P/S)} & Avg\\
\cmidrule(lr){2-11}
\rowcolor{ourback}& \method\ (all at once) & \textbf{90.10} & \textbf{88.83} & \textbf{94.58} & \textbf{95.99} & \textbf{70.23} & \textbf{89.17} & \textbf{91.63} & \textbf{92.00} & \textbf{89.07}\\
& \method\ (incremental, multi-task base) & \underline{87.63} & \underline{85.34} & \underline{86.80} & \underline{92.89} & 45.06 & \underline{79.78} & 83.81 & 87.68 & \underline{81.13}\\
& \method\ (incremental, single-task base) & 82.67 & 82.06 & 84.50 & 88.76 & 49.87 & 77.26 & \underline{84.74} & 86.45 & 79.54\\
& Sequential adapter fine-tuning (baseline) & 34.17 & 31.57 & 64.18 & 75.23 & \underline{63.02} & 46.93 & 82.02 & \underline{91.76} & 61.11\\
\multirow{-5}{*}{\rotatebox[origin=c]{90}{\tiny\textbf{GLUE}}}& Sequential full fine-tuning (baseline) & 47.01 & 79.53 & 79.28 & 69.72 & 56.52 & 53.79 & 78.93 & 91.62 & 69.55\\
\midrule
& Method & BoolQ {\tiny(acc)} & CB {\tiny(acc/F1)} & RTE {\tiny(acc)} & WiC {\tiny(acc)} & WSC {\tiny(acc)} & COPA {\tiny(acc)} & MultiRC {\tiny(F1a/EM)} & ReCoRD {\tiny(EM/F1)} & Avg\\
\cmidrule(lr){2-11}
\rowcolor{ourback}& \method\ (all at once) & \textbf{84.86} & \underline{91.09} & \textbf{85.92} & \textbf{69.28} & \underline{87.50} & \textbf{91.00} & \underline{59.84} & \textbf{87.02} & \textbf{82.06}\\
& \method\ (incremental, multi-task base) & \underline{84.19} & \textbf{91.12} & \underline{84.84} & \underline{68.97} & \textbf{88.46} & 86.00 & 56.66 & 85.50 & \underline{80.72}\\
& \method\ (incremental, single-task base) & 68.10 & 86.86 & 64.26 & 62.70 & 76.92 & 86.00 & \textbf{60.10} & 84.57 & 73.69\\
& Sequential adapter fine-tuning (baseline) & 59.88 & 71.77 & 71.48 & 58.62 & 61.54 & \underline{89.00} & 48.90 & \underline{85.67} & 68.36\\
\multirow{-5}{*}{\rotatebox[origin=c]{90}{\tiny\textbf{SuperGLUE}}}& Sequential full fine-tuning (baseline) & 65.26 & 72.84 & 63.90 & 55.17 & 61.54 & 78.00 & 45.35 & 82.76 & 65.60\\
\bottomrule
\end{tabular}}
\end{table}

\subsection{Importance-based vs.\ permutation-test rank estimation}\label{sec:apprankcompare}
Both estimators allocate rank per structure; they differ only in how each structure's rank is chosen.
We compare the default importance-based allocator (Appendix~\ref{sec:apprank}), which prunes singular
values under a decaying budget, against the budget-free permutation-test estimator on multi-task GLUE
and SuperGLUE under the same backbone and schedule, both at matched effective per-task rank $8$. On
SuperGLUE the permutation test reaches the best average
($82.06$ vs.\ $81.21$; Table~\ref{tab:rank_est}), winning CB, WSC, COPA, and ReCoRD while the allocator
keeps BoolQ, RTE, WiC, and MultiRC. On GLUE the picture reverses, and the importance-based allocator holds a
small edge ($89.07$ vs.\ $88.54$), keeping SST-2, CoLA, RTE, and STS-B while the permutation test takes
MNLI, QQP, QNLI, and MRPC. Neither estimator dominates across benchmarks; since the permutation test
removes the rank-budget hyperparameter, it is the more convenient default wherever it stays
competitive.

\begin{table}[h]\centering\scriptsize
\caption{Importance allocator vs.\ budget-free permutation-rank test at matched per-task rank $8$.
The permutation test wins on SuperGLUE; the importance allocator edges it on GLUE.}
\label{tab:rank_est}
\renewcommand{\arraystretch}{1.5}
\resizebox{\textwidth}{!}{%
\begin{tabular}{clccccccccc}
\toprule
& Rank-estimation strategy & MNLI {\tiny(acc)} & QQP {\tiny(acc/F1)} & QNLI {\tiny(acc)} & SST-2 {\tiny(acc)} & CoLA {\tiny(Mcc)} & RTE {\tiny(acc)} & MRPC {\tiny(acc/F1)} & STS-B {\tiny(P/S)} & Avg\\
\cmidrule(lr){2-11}
& Importance-based (eff-rank $8$) & \underline{90.10} & \underline{90.34}/\underline{87.32} & \underline{94.58} & \textbf{95.99} & \textbf{70.23} & \textbf{89.17} & \underline{90.20}/\underline{93.06} & \textbf{92.03}/\textbf{91.98} & \textbf{89.07}\\
\multirow{-3}{*}{\rotatebox[origin=c]{90}{\tiny\textbf{GLUE}}}& Permutation-test (eff-rank $8$) & \textbf{90.25} & \textbf{90.39}/\textbf{87.39} & \textbf{94.60} & \underline{95.18} & \underline{67.97} & \underline{87.73} & \textbf{90.69}/\textbf{93.33} & \underline{91.72}/\underline{91.59} & \underline{88.54}\\
\midrule
& Rank-estimation strategy & BoolQ {\tiny(acc)} & CB {\tiny(acc/F1)} & RTE {\tiny(acc)} & WiC {\tiny(acc)} & WSC {\tiny(acc)} & COPA {\tiny(acc)} & MultiRC {\tiny(F1a/EM)} & ReCoRD {\tiny(EM/F1)} & Avg\\
\cmidrule(lr){2-11}
& Importance-based (eff-rank $8$) & \textbf{85.87} & \textbf{92.86}/\underline{86.21} & \textbf{88.09} & \textbf{69.91} & \underline{78.85} & \underline{90.00} & \textbf{78.43}/\textbf{44.07} & \underline{85.91}/\underline{86.48} & \underline{81.21}\\
\multirow{-3}{*}{\rotatebox[origin=c]{90}{\tiny\textbf{SuperGLUE}}}& Permutation-test (eff-rank $8$) & \underline{84.86} & \textbf{92.86}/\textbf{89.32} & \underline{85.92} & \underline{69.28} & \textbf{87.50} & \textbf{91.00} & \underline{77.28}/\underline{42.39} & \textbf{86.76}/\textbf{87.28} & \textbf{82.06}\\
\bottomrule
\end{tabular}}
\end{table}

\subsection{Low-rank structure of the task adaptation}\label{sec:appeffrank}
\begin{figure}[h]\centering
\includegraphics[width=0.68\textwidth]{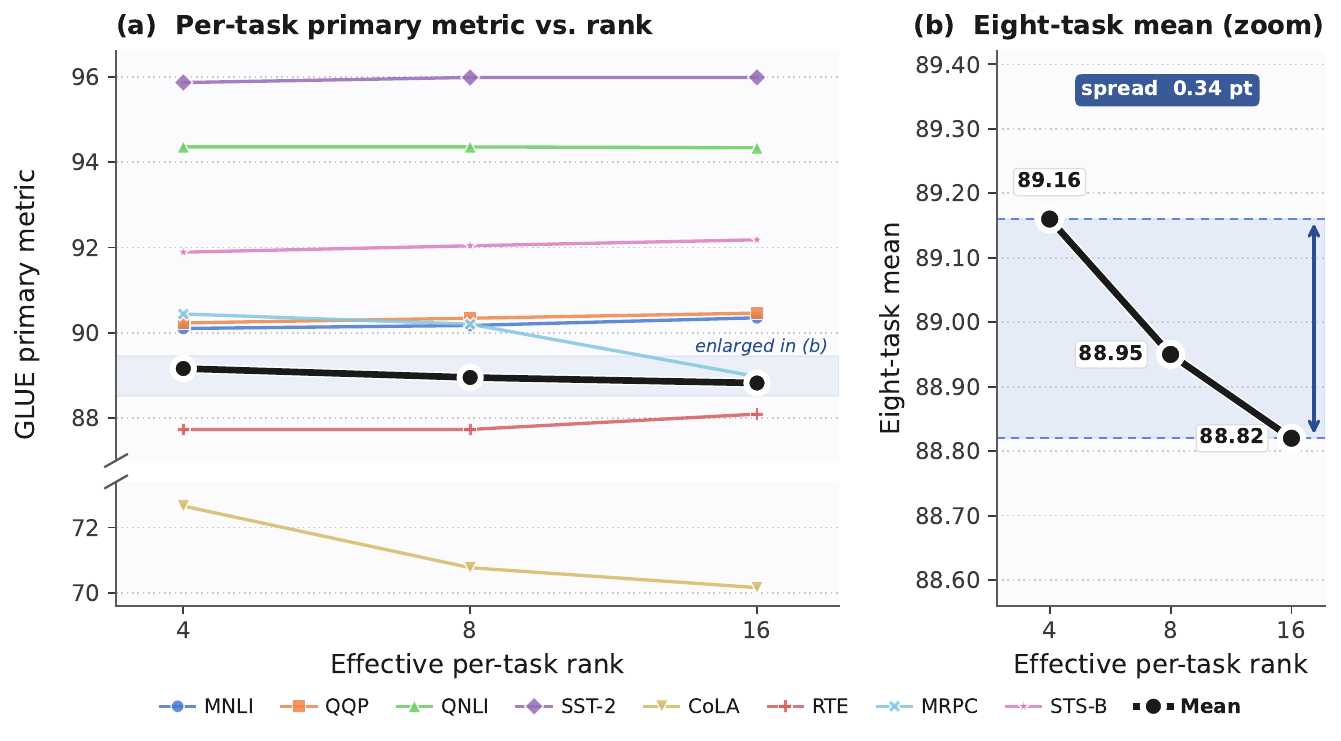}
\caption{The task adaptation \method\ learns is low rank on multi-task GLUE (DeBERTaV3-base).
(a) per-task primary metric and eight-task mean at rank $4$, $8$, and $16$; (b) the mean zoomed,
varying by only $0.34$ points across the fourfold range.}
\label{fig:effrank}
\end{figure}

The effective per-task rank is the number of adapted directions each task receives once its shared
Joint and own Individual are folded together. We claim this adaptation is intrinsically \emph{low
rank}. The Joint plus Individual split concentrates the task signal in a few directions. We test this on the eight
GLUE tasks (DeBERTaV3-base) at effective per-task rank $4$, $8$, and $16$ (target rank $2$, $4$, $8$
per pool), fixing the backbone, schedule ($20{,}000$ steps), and seed. Figure~\ref{fig:effrank}
reports the primary GLUE metric per task (Matthews for CoLA, Pearson for STS-B, accuracy otherwise)
and the eight-task mean.

The mean is almost flat across a fourfold change in budget, moving only $0.34$ points ($89.16$,
$88.95$, $88.82$); the smallest budget is in fact the strongest. The Joint already spans the
directions shared by all tasks and each Individual adds only the few residual directions its task
needs, so rank $4$ already covers the subspace of $\Delta\Wm_t=\Jm+\Im_t$ and any extra rank only
fits near-zero-singular-value components that carry noise, not signal. The per-task split confirms
this. Large-data tasks (MNLI, QQP, SST-2, STS-B) gain slightly with more rank, while small,
high-variance tasks overfit the noisy directions and regress (CoLA $72.66$ to $70.16$, MRPC $90.44$
to $88.97$), pulling the mean down.

\subsection{Interpretability Analysis}\label{sec:appexp}
To understand why the Joint/Individual split drives our main benchmark results, we analyze the
structure it learns. We expect the Joint to capture cross-task shared structure, each Individual
to capture task-specific structure, and the two to remain orthogonal and interpretable. We cast
these expectations as testable hypotheses and curate a targeted experiment for each.

\subsubsection{Hypotheses}
Our experiments test the following hypotheses.
\begin{description}[leftmargin=0pt,style=unboxed]
  \item[H1: Separation.] The Individual subspaces are diverse and each $\Im_t$ is
  near-orthogonal to $\Jm$.
  \item[H2: Task geometry.] Pairwise similarity of the task signatures is structured rather than
  uniform and gives each held-out task a consistent routing target.
  \item[H3: Rank-budget split between structures.] How the fixed rank budget is divided between the
  shared Joint and the per-task Individuals affects multi-task performance, and we test which split
  works best.
\end{description}

\subsubsection{Experiments}
\begin{figure}[h]\centering
\includegraphics[width=\linewidth]{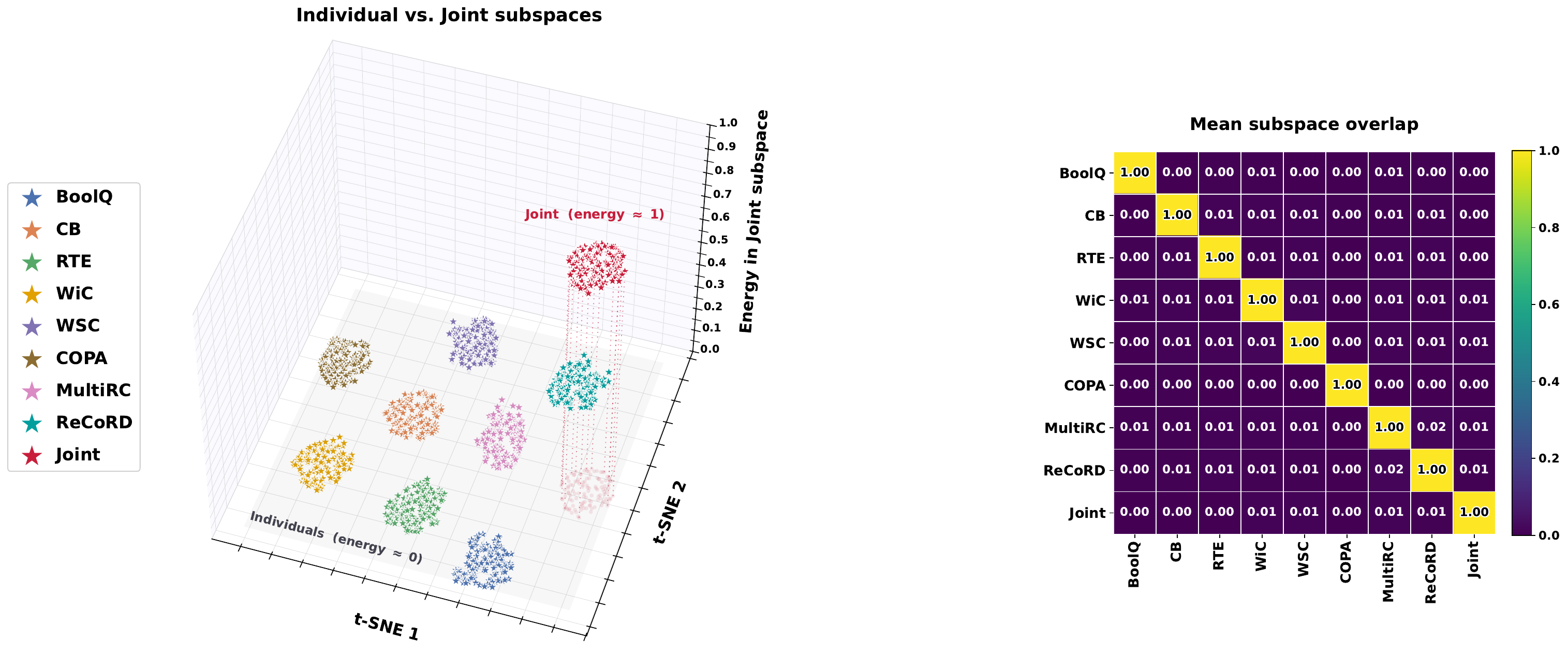}
\caption{Learned Individual and Joint subspace geometry.}
\label{fig:tsne_indiv}
\end{figure}

\paragraph{E1: Learned subspace geometry (tests H1).}
To examine how the learned subspaces span the hidden space relative to one another, we sample unit vectors from each
task's aggregate Individual subspace and from the shared Joint subspace and visualize them in
Figure~\ref{fig:tsne_indiv}. The horizontal plane is a t-SNE of these vectors, and the
vertical axis is each vector's energy in the Joint subspace,
\[
  z(v) \;=\; \lVert P_{\Jm}\, v \rVert^2 \;=\; \sum_{\ell} \lVert U_\ell^{\top} v_\ell \rVert^2 \;\in\; [0,1],
\]
where $P_{\Jm}$ is the orthogonal projector onto the Joint column space (assembled per module
$\ell$ from an orthonormal Joint basis $U_\ell$) and $v$ is unit norm. So $z$ is the fraction of a
vector's squared length that lies inside the Joint subspace; $z\!\approx\!1$ means the vector lives
almost entirely in the Joint, and $z\!\approx\!0$ means it is orthogonal to the Joint. The companion
matrix on the right reports the pairwise overlaps between the per-task Individual bases and the
Joint.

As shown in figures, the eight Individual subspaces settle at low energy in the Joint direction while the Joint
vectors rise to one, so every Individual carries almost no energy in the Joint; that is, each
$\Im_t$ is near-orthogonal to $\Jm$ as H1 predicts. The companion matrix confirms this, since the
Individual-in-Joint entries are near zero (the diagonal, each subspace against itself, is one by
construction). The Individuals also occupy distinctly separated regions,
so their cross-task overlaps are empirically small; this separation is learned, not imposed, since
only the Individual\,$\perp$\,Joint constraint is enforced during training, yet it yields
non-overlapping learned subspaces alongside the shared Joint structure.

\begin{figure}[t]\centering
\begin{minipage}[b]{0.45\textwidth}\centering
\includegraphics[width=\textwidth]{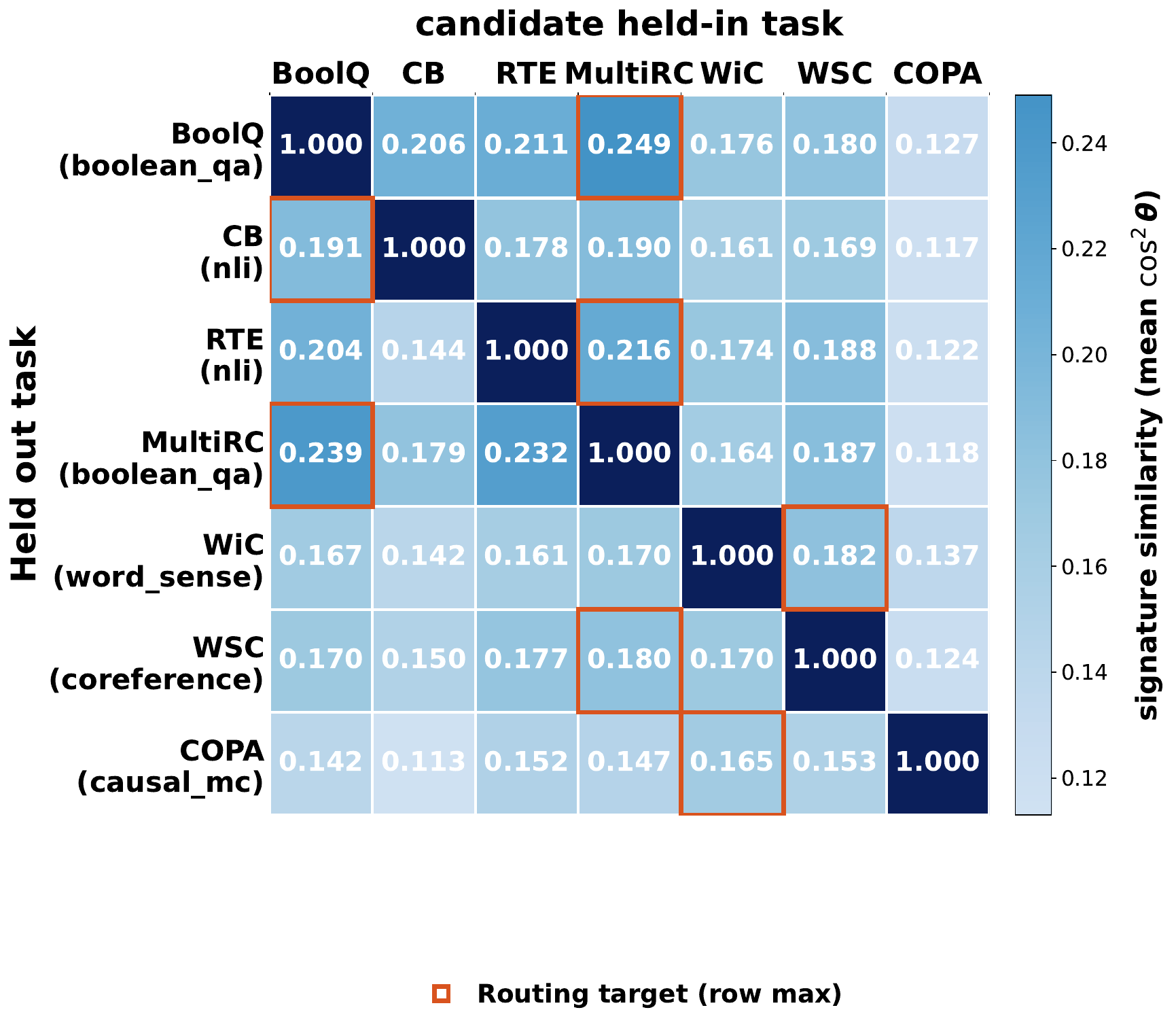}\\[2pt]
{\footnotesize (a) SuperGLUE cross-task pool}
\end{minipage}\hfill%
\begin{minipage}[b]{0.45\textwidth}\centering
\includegraphics[width=\textwidth]{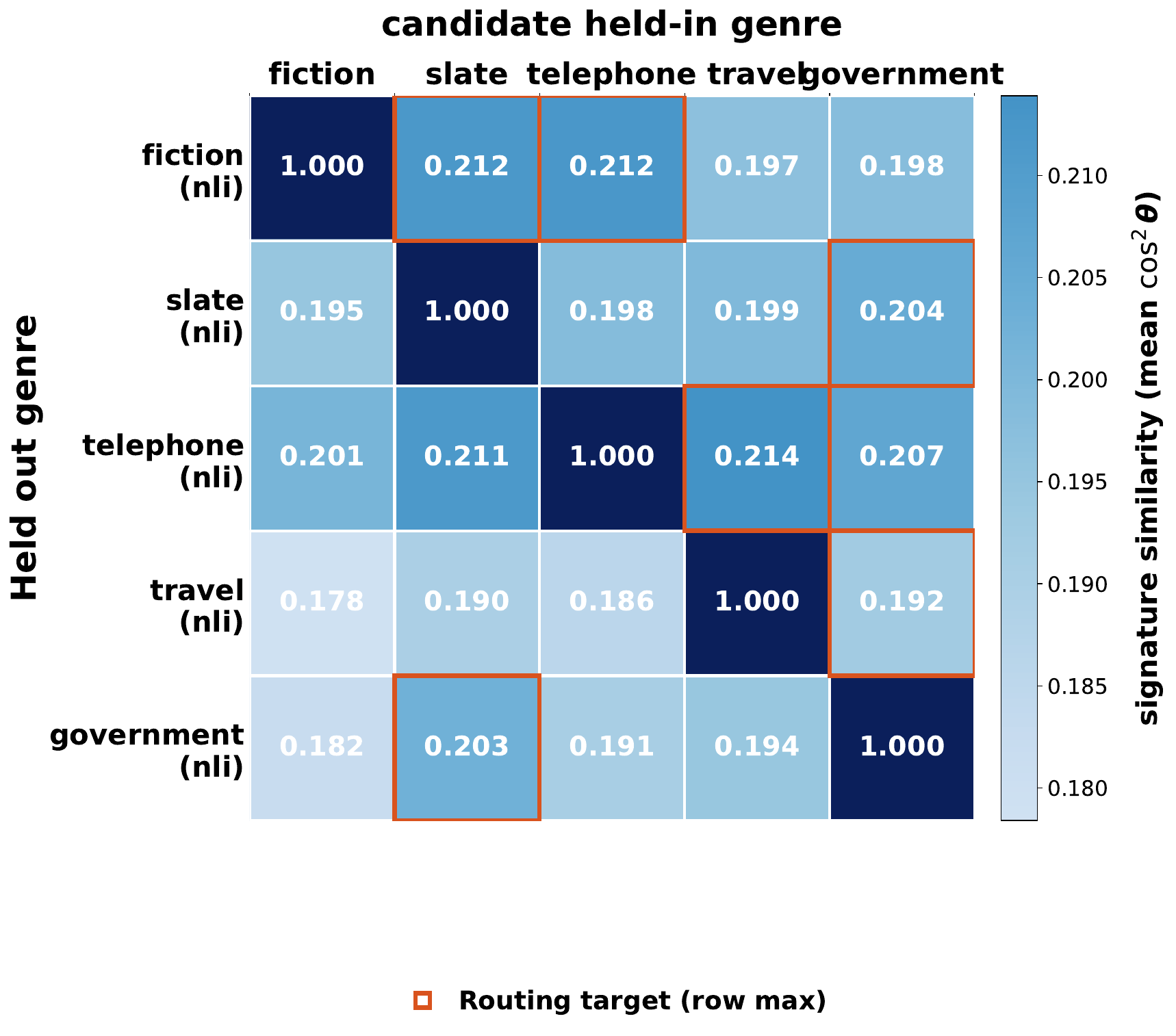}\\[2pt]
{\footnotesize (b) MNLI cross-genre pool}
\end{minipage}%
\caption{Held-out task signature similarity to the held-in Individuals; each row's boxed cell is its routing target.}
\label{fig:reuse_overlap}
\end{figure}

\paragraph{E2: Task signature similarity and routing (tests H2).}
A held-out task is never used during structure learning, so the learned arrangement of the
trained Individuals does not bind it, and its reference signature is free to align with
related held-in Individuals. Figure~\ref{fig:reuse_overlap} reports that principal-angle
similarity, and each held-out task routes to its highest-similarity held-in neighbor (boxed).
The boolean\_qa pair (BoolQ, MultiRC) is mutually highest; CB routes to BoolQ and RTE to
MultiRC, while WiC, WSC, and COPA route to WSC, MultiRC, and WiC.

\vspace{-1.2em}

\begin{figure}[H]\centering
\includegraphics[width=0.68\textwidth]{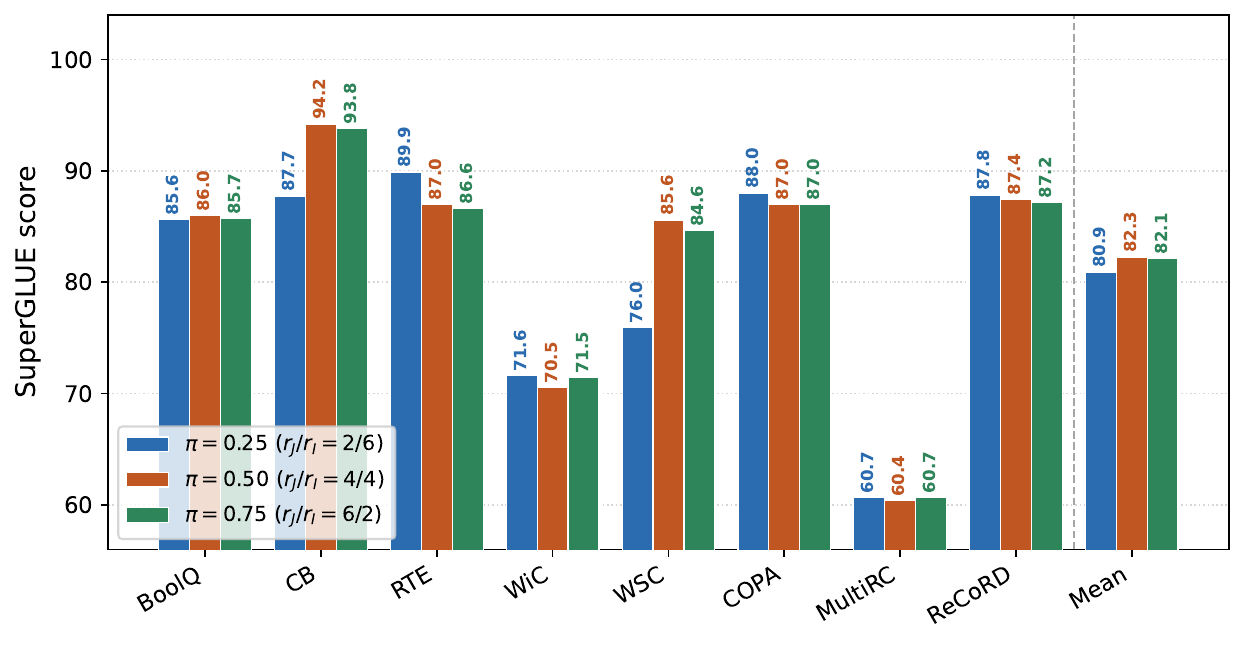}
\caption{Rank-budget split between the shared Joint and the per-task Individuals, swept on multi-task SuperGLUE at fixed per-task effective rank $8$.}
\label{fig:sharing}
\end{figure}

\paragraph{E3: Rank-budget split between structures (tests H3).}
At a \emph{fixed} per-task budget of effective rank $8$, we split that budget between the shared
Joint and the per-task Individual and vary the sharing ratio $\pi=r_J/(r_J+r_I)$, training \method\
multi-task on the eight SuperGLUE tasks at each split and recording the multi-task average
(Figure~\ref{fig:sharing}). The allocation ratio has a measurable effect on performance, and a balanced split is optimal. The even $4/4$
split ($\pi{=}0.50$) attains the highest average at $82.26$, above both the Individual-heavy $2/6$ split
($\pi{=}0.25$, $80.92$) and the Joint-heavy $6/2$ split ($\pi{=}0.75$, $82.14$). Giving the Joint too
little ($\pi{=}0.25$) leaves cross-task structure underused, whereas giving it too much
($\pi{=}0.75$) starves the Individuals and degrades the more task-specific tasks such as WSC
($85.58\!\to\!84.62$) and RTE ($87.00\!\to\!86.64$). The two
extremes recover baselines we already report; $\pi{=}0$ is Individuals only ($T$ independent LoRAs
with no transfer) and $\pi{=}1$ is a single fully-shared adapter (the LoRA/AdaLoRA MTL setting of
Table~\ref{tab:multi}), so the interior points are where the Joint plus Individual split pays off.
This best fixed split ($82.26$) slightly exceeds the adaptive rank-estimation allocator reported in
Tables~\ref{tab:multi} and~\ref{tab:rank_est} ($82.06$ on SuperGLUE); the allocator sets each
structure's rank automatically without an explicit search over $\pi$, staying within a fraction of a point of the
best hand-tuned split while removing that search. The allocator therefore automates what is
otherwise a manual rank-allocation procedure, resolving the trade-off between an expensive grid
search and accuracy at no measurable cost in performance.

\end{document}